\pdfoutput=1

\documentclass[11pt]{article}

\usepackage[final]{acl}

\usepackage{times}
\usepackage{subcaption}
\usepackage{placeins}
\usepackage{latexsym}
\usepackage{CJKutf8}
\usepackage{amsmath}
\usepackage{float}
\usepackage{amsfonts}
\usepackage[T1]{fontenc}

\usepackage[utf8]{inputenc}

\usepackage{microtype}

\usepackage{inconsolata}

\usepackage{graphicx}

\title{Causal Language Control in Multilingual Transformers via Sparse Feature Steering}

\author{
 \textbf{Cheng-Ting Chou\textsuperscript{1}\thanks{Lead Author}},
 \textbf{George Liu\textsuperscript{2}},
 \textbf{Jessica Sun\textsuperscript{3}},
 \textbf{Cole Blondin\textsuperscript{4}},
\\
 \textbf{Kevin Zhu\textsuperscript{4}},
 \textbf{Vasu Sharma\textsuperscript{5}},
 \textbf{Sean O'Brien\textsuperscript{4}},
\\
\\
 \textsuperscript{1}University of Illinois Urbana-Champaign \quad
 \textsuperscript{2}University of Maryland, College Park\\
 \textsuperscript{3}Barnard College \quad
 \textsuperscript{4}Algoverse AI Research \quad
 \textsuperscript{5}Meta FAIR Lab
\\
 \small{
   \textbf{Correspondence:} \href{mailto:ctchou3@illinois.edu}{ctchou3@illinois.edu}
 }
}

\begin{document}
\maketitle
\begin{abstract}
Deterministically controlling the target generation language of large multilingual language models (LLMs) remains a fundamental challenge, particularly in zero-shot settings where neither explicit language prompts nor fine-tuning are available. In this work, we investigate whether sparse autoencoder (SAE) features, previously shown to correlate with interpretable model behaviors, can be leveraged to steer the generated language of LLMs during inference. Leveraging pretrained SAEs on the residual streams of Gemma-2B and Gemma-9B, we identify features whose activations differ most significantly between English and four target languages: Chinese, Japanese, Spanish, and French. By modifying just a single SAE feature at one transformer layer, we achieve controlled language shifts with up to 90\% success, as measured by FastText language classification, while preserving semantic fidelity according to LaBSE (Language-Agnostic BERT Sentence Embedding) similarity. Our analysis reveals that language steering is most effective in mid-to-late transformer layers and is amplified by specific attention heads disproportionately associated with language-sensitive SAE features. These results demonstrate the promise of sparse feature steering as a lightweight and interpretable mechanism for controllable multilingual generation.
\end{abstract}

\section{Introduction}

Large language models (LLMs) trained on multilingual corpora have achieved remarkable success in generating and understanding text across diverse languages. However, these models often entangle abstract representations within shared internal structures, making it difficult to isolate or manipulate specific language behavior \cite{elhage2022superposition}. 

Recent work on sparse autoencoders (SAEs) has shown that transformer residual streams can be decomposed into interpretable, behavior-aligned directions \cite{bricken2023monosemanticity, huben24}. Building on this insight, methods such as Feature-Guided Activation Additions (FGAA) \cite{soo2025interpretable} and SAE-Targeted Steering (SAE-TS) \cite{chalnev2024saets} demonstrate that selectively activating SAE features can causally steer LLM behavior. These techniques focus on improving the precision and coherence of general-purpose steering, but their application to controlling output \textbf{language identity} remains unexplored.

To investigate this, we adapt \textbf{SAE feature steering}, a simple method for manipulating the language of multilingual model outputs by modifying one SAE feature activation at inference time \cite{bricken2023monosemanticity}. As shown in Table~\ref{tab:steering_examples}, steering just one optimal feature can shift output language from English to Chinese, Japanese, Spanish, or French without using prompts, labels, or fine-tuning.

Our approach leverages pretrained SAEs from the Gemma-Scope project to identify language-specific features that differentiate English from four target languages (French, Spanish, Chinese, and Japanese) \cite{lieberum2024gemmascopeopensparse}. By shifting selected features at a single transformer layer, we induce high-quality generation in the target language while preserving semantic content.

\begin{table*}[t]
  \centering
  \begin{tabular}{p{3cm} | p{12cm}}
    \hline
    \textbf{Prompt} & \texttt{Artificial intelligence is important because} \\
    \hline
    \textbf{No Steer} &
    Artificial intelligence is important because it can help us solve problems that we would not be able to solve otherwise. For example, AI can help us find new ways to cure diseases or make better decisions about how to allocate resources. Artificial intelligence is a branch of computer science that deals with the simulation of human intelligence in machines. \\
    \hline
    \textbf{Steered (Chinese)} &
    \begin{CJK}{UTF8}{gbsn}Artificial intelligence is important because它帮助我们工作更快，更有效地达到我们的目标。AI可以解决人类无法处理的大量数据，并通过其学习算法提高效率。它使我们能够更快、更准确地做出决定，并提高整体生产力。人工智能的未来充满希望。\end{CJK} \\
    \hline
    \textbf{Steered (Spanish)} &
    Artificial intelligence is important because es la inteligencia que tiene el ser humano. El ser humano tiene una inteligencia que no puede tener ningún otro animal, porque no puede pensar como un ser humano. En este sentido, tenemos que hablar de la inteligencia artificial o IA. \\
    \hline
    \textbf{Steered (Japanese)} &
    \begin{CJK}{UTF8}{min}Artificial intelligence is important because人間はそれを活用することで、様々な問題を解決することが可能になるからです。例えば、医療分野ではAIが患者さんの状態を分析して適切な治療方法を提案したり、交通事故の原因を推定したりするなど、様々な用途に使われています。\end{CJK} \\
    \hline
    \textbf{Steered (French)} &
    Artificial intelligence is important because c'est le futur. La société est en train de passer au numérique, et les entreprises doivent être préparées à suivre. Le problème est que le numérique est une nouvelle technologie, ce qui signifie que les gens ne savent pas forcément comment s'en servir. \\
    \hline
  \end{tabular}
  \caption{
    \textbf{Examples of multilingual language steering.}
    We show Gemma-2-9B's continuation for a single prompt with and without SAE feature steering. The steered outputs exhibit clear and fluent shifts into the target languages, while preserving semantic relevance to the original prompt.
    }
  \label{tab:steering_examples}
\end{table*}

\paragraph{Key Contributions}
\begin{itemize}
    \item \textbf{Causal language control via sparse feature steering}: We introduce a method for shifting LLM output language by editing a small set of language-specific SAE features at inference time without retraining or prompt engineering. Our method is closely related to FGAA and SAE-TS but uniquely focuses on language identity control.

    \item \textbf{Evaluation across four target languages}: We test our method on translations from English to French, Spanish, Chinese, and Japanese, using both semantic similarity (LaBSE) and automatic language identification (FastText) \cite{feng-etal-2022-language, joulin2016bag}. Steering just one feature per language yields target-language outputs in up to 90\% of cases.

    \item \textbf{Layerwise analysis of steering effects}: In contrast to prior work that applies SAE steering at a fixed layer, we perform a comprehensive analysis across all transformer layers, revealing where language features are most causally effective and how steering strength evolves by depth.

    \item \textbf{Interpretability through attention and residual decomposition}: We trace the origin of language-aligned SAE features across layers and show that language steering is amplified by specific attention heads, revealing architectural mechanisms that support steerable behavior.
\end{itemize}

\section{Related Works}

\paragraph{Sparse Autoencoders and Mechanistic Interpretability}
Large language models often exhibit superposition, where neurons encode multiple unrelated features due to capacity constraints \cite{elhage2022superposition}. Sparse autoencoders (SAEs) address this by learning a sparse basis over model activations that decomposes them into more interpretable, monosemantic features \cite{bricken2023monosemanticity, huben24}. \citet{bricken2023monosemanticity} demonstrated that these features often align with human-understandable patterns—e.g., legal or multilingual text—and can be used to analyze model behavior more precisely than neuron-level inspection.

\paragraph{Activation Steering with Sparse Features}
Recent work has shown that SAE features are not only interpretable, but causally manipulable. \textbf{FGAA} \cite{soo2025interpretable} demonstrated that activating a small set of SAE features can steer model behavior in interpretable directions. \textbf{SAE-TS} \cite{chalnev2024saets} refined this by constructing steering vectors that target specific features while suppressing others to reduce side effects. However, these methods evaluate steering at a single mid-layer (e.g., layer 12), leaving open how steerability varies across layers. Our work performs layerwise steering and analysis, revealing how causal control of language behavior shifts with depth.

\paragraph{Controlling Multilingual Behavior}
Multilingual LLMs implicitly encode language identity, and recent work shows it can be causally manipulated. \citet{tang-etal-2024-language} identified language-specific neurons in BLOOM and LLaMA-2 whose activation controlled output language. Similarly, \citet{chang-etal-2022-geometry} showed that shifting hidden states along language directions in XLM-R could toggle model outputs. Unlike these neuron-level or architectural interventions, our work steers multilingual behavior using sparse, interpretable features—requiring no retraining or architectural changes.

\section{Methods}

Our methodology comprises of four key components: (1) we identified language-representative features by analyzing sparse autoencoder (SAE) activations across different languages; (2) we selected SAE features that exhibited significant activation differences between English and other target languages; (3) we applied feature steering by intervening on these features during model inference to test their causal influence; and (4) we analyzed how language-specific features were distributed and propagated across transformer layers. We detailed each of these components below.

\subsection{Model Architectures (Gemma-2-2B and Gemma-2-9B)}

We used two decoder-only transformer language models, Gemma-2-2B and Gemma-2-9B, with 26 and 42 layers respectively \cite{gemma_2024}. These models were selected because pretrained sparse autoencoders (SAEs) from the Gemma-Scope project \cite{lieberum2024gemmascopeopensparse} are readily available for this model family, enabling efficient and interpretable analysis. Both models followed a GPT-style architecture and were trained on a multilingual corpus containing English, French, Spanish, Chinese, and Japanese, using a shared byte-pair encoding (BPE) vocabulary. Without explicit language supervision or fine-tuning, these models learned internal representations shaped by the statistical patterns of each language. To study how individual languages were encoded, we input parallel sentences in English and each of the four target languages, and extracted the hidden activations $\mathbf{h}^l_{\text{lang}, i}$ at each layer $l$ and token position $i$. These activations served as input for our SAE-based feature analysis.

\subsection{Sparse Autoencoder Design and Training}

To analyze internal representations, we used the pretrained sparse autoencoders (SAEs) from the Gemma-Scope project \cite{lieberum2024gemmascopeopensparse}, which provided a compressed and interpretable basis over each model layer’s residual stream. For each layer $l$, the SAE encoded the residual activation $\mathbf{h}^l \in \mathbb{R}^d$ into a sparse code $\mathbf{z}^l \in \mathbb{R}^{16384}$ via a linear encoder, such that $\mathbf{h}^l \approx D^l(\mathbf{z}^l)$, where $D^l$ was the linear decoder. The residual dimension $d$ was 2304 for Gemma-2-2B and 3584 for Gemma-2-9B. Each dimension of $\mathbf{z}^l$ corresponded to a learned feature that was active only for a small subset of inputs, yielding a sparse and interpretable decomposition. We referred to the nonzero components (without additional thresholding) of $\mathbf{z}^l$ as the active SAE features, which contained the basis for identifying language-indicative directions in representation space.

\subsection{Identifying Language-Specific Features}
\label{subsec:identify-language-features}

To identify features associated with specific languages, we computed activation differences between English sentences and their corresponding translations in a target language. We used 1,000 parallel sentence pairs per language from the Tatoeba Project (via ManyThings.org) \cite{tatoeba, manythings}. Our method followed the general contrastive framework introduced in FGAA \cite{soo2025interpretable}, but adapted it to identify language-divergent SAE features using parallel sentence corpora. Rather than optimizing a steering vector, we used the contrastive signal to select individual features for direct activation-based intervention. For each transformer layer $l$, we extracted the residual stream activations $\mathbf{h}^l(x)$ for input $x$ and applied the SAE encoder $f$ to obtain sparse feature activations $\mathbf{z}^l(x) = f(\mathbf{h}^l(x))$.

Let $X^{\text{EN}}$ and $X^{\text{TL}}$ denote the sets of English and target language inputs (e.g., French, Spanish), respectively. We computed a contrastive feature difference vector for each layer $l$ as:

\begin{align*}
\Delta^l 
&= \frac{1}{|X^{\text{TL}}|} \sum_{x \in X^{\text{TL}}} \bar{f}(\mathbf{h}^l(x)) \\
&\quad - \frac{1}{|X^{\text{EN}}|} \sum_{x \in X^{\text{EN}}} \bar{f}(\mathbf{h}^l(x))
\end{align*}

where $\bar{f}(\mathbf{h}^l(x))$ denotes the \textit{mean SAE feature activation across all tokens} in input $x$ at layer $l$. We also defined an alternative variant using only the activation at the final token $\texttt{EOS}(x)$:

\begin{align*}
\Delta^l_{\text{final}} &= \frac{1}{|X^{\text{TL}}|} \sum_{x \in X^{\text{TL}}} f(\mathbf{h}^l(x)_{-1}) \\
&\quad - \frac{1}{|X^{\text{EN}}|} \sum_{x \in X^{\text{EN}}} f(\mathbf{h}^l(x)_{-1})
\end{align*}

We computed both $\Delta^l$ and $\Delta^l_{\text{final}}$ for all four language pairs and all layers. The top-$k$ features with the largest absolute difference in each direction were selected for steering. Unless otherwise stated, we fixed $k=3$ for simplicity and interpretability. With 1,000 translated sentence pairs per language, the features exhibiting the largest mean activation differences are most likely to encode core linguistic distinctions rather than incidental or semantic variation. Selecting $k=3$ features ensures coverage of the dominant language-indicative signals.

\subsection{Feature Steering Method}

To test whether these features causally influence language generation, we performed targeted feature interventions at each transformer layer. Given an English input $x$, we extracted the residual stream activations $\mathbf{h}^l(x)$ and computed its sparse code $\mathbf{z}^l = f(\mathbf{h}^l(x))$. For each selected feature $j$, we applied the corresponding language activation offset $\Delta^l_j$ (or $\Delta^l_{\text{final},j}$) derived from the training set:

\[
z'^l_j = z^l_j + \Delta^l_j
\]

The modified latent code $\mathbf{z}'^l$ is decoded via the SAE decoder $g$ to obtain the updated residual activation:

\[
\mathbf{h}'^l = \mathbf{h}^l + g(\mathbf{z}'^l - \mathbf{z}^l)
\]

This intervened representation $\mathbf{h}'^l$ replaced the original hidden state at layer $l$, and the model completed the forward pass using the modified activation. All generations were produced using a decoding temperature of 0.5 and a maximum output length of 50 tokens. This approach resembled prior work on activation steering \cite{soo2025interpretable, chalnev2024saets}, but is applied to language identity features and extended across all layers. By comparing completions before and after intervention, we assessed whether activating these features reliably steers output toward the target language.

\subsection{Evaluation Metric for Outputs} 
\label{Evaluation Metrics for Output Quality}
To evaluate the effectiveness of feature steering, we used a separate set of 500 English prompts derived from the Dolly-15k dataset \cite{DatabricksBlog2023DollyV2}. These prompts span diverse domains (e.g., science, food, personal interests) and were manually proofread to ensure open-endedness and semantic clarity. Although the generated answers to these prompts may vary, the structured and semantically rich nature of the prompts helps constrain the output to remain within the intended domain, enabling meaningful comparisons post-steering. Each generated output is assessed using two metrics: linguistic match and semantic preservation.

\begin{itemize}
\item \textbf{Linguistic match} is determined using FastText, a lightweight and accurate language identifier trained on over 170 languages \cite{joulin2016bag}. Each steered sentence is assigned a score of 1 if classified as the target language, and 0 otherwise.
\item \textbf{Semantic preservation} is measured via cosine similarity between sentence embeddings produced by LaBSE, a multilingual encoder trained for cross-lingual retrieval \cite{feng-etal-2022-language}. This reflected the semantic similarity between the original English input and the steered output. To enforce fidelity to both form and meaning, outputs that failed the linguistic match check are assigned a semantic score of 0.
\end{itemize}

\begin{figure*}[t]
  \centering
  \begin{minipage}[t]{0.49\linewidth}
    \centering
    \includegraphics[width=\linewidth]{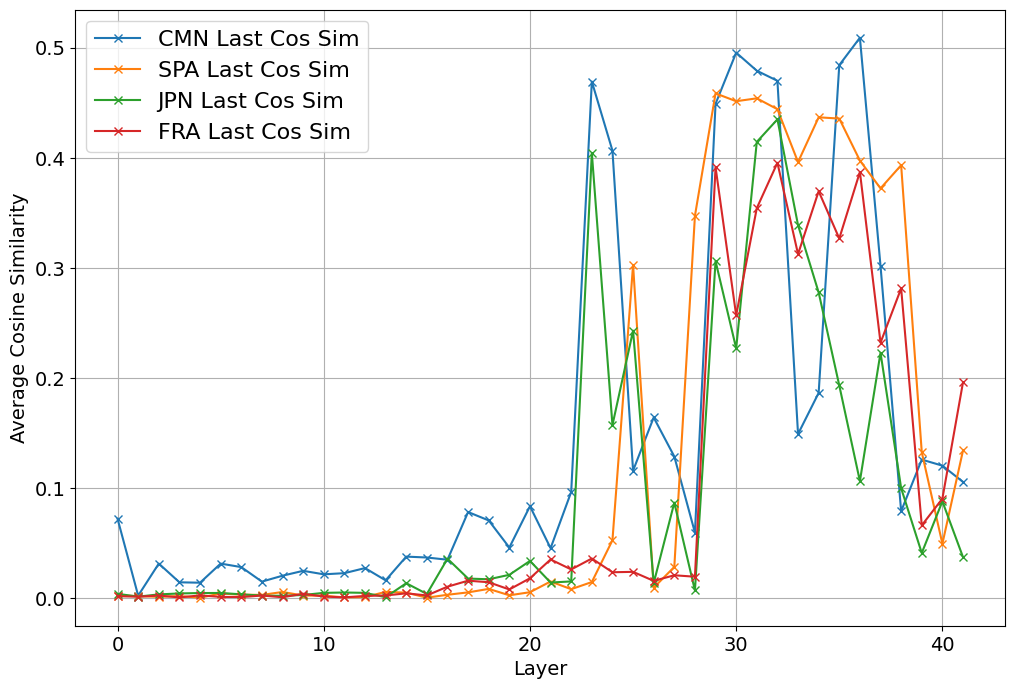}
    \caption{LaBSE semantic similarity scores for steered outputs across Gemma-2-9B layers, using last-token-selected language features for Chinese (CMN), Spanish (SPA), Japanese (JPN), and French (FRA). The results show that steering effectiveness varies across layers, with peak semantic alignment occurring at mid to late layers for different languages.}
    \label{fig:9b_last_sims}
  \end{minipage}%
  \hfill
  \begin{minipage}[t]{0.49\linewidth}
    \centering
    \includegraphics[width=\linewidth]{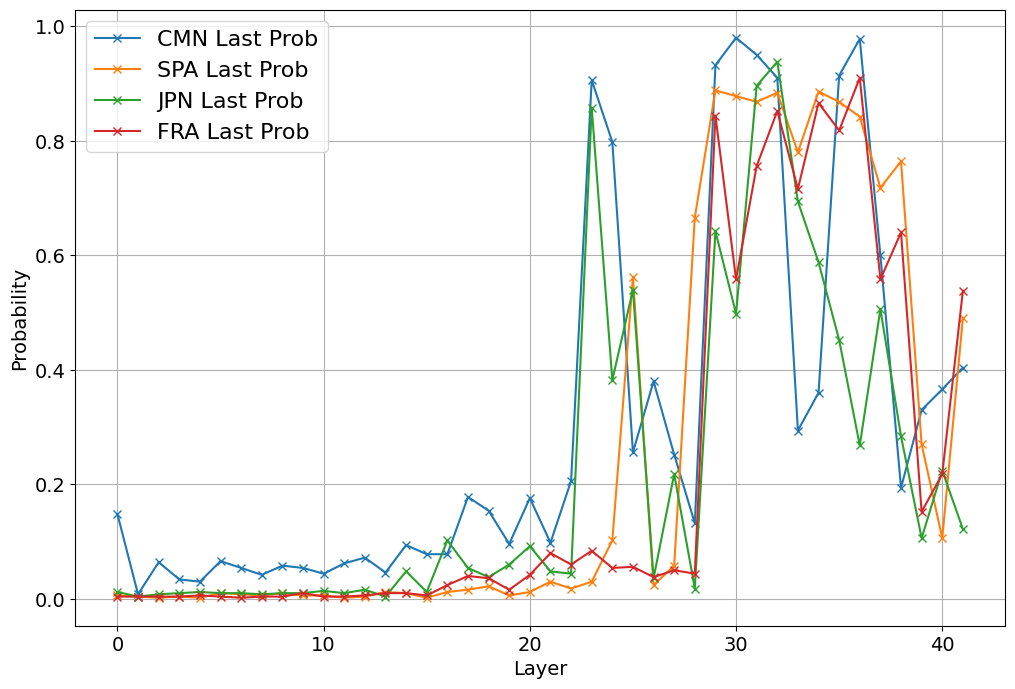}
    \caption{FastText classification probabilities of the same steered outputs, revealing layer-specific differences in how strongly outputs reflect the target language. Later layers generally show higher classification confidence, indicating greater controllability through steering at those depths.}
    \label{fig:9b_last_probs}
  \end{minipage}
\end{figure*}

\subsection{Layerwise and Behavioral Analysis of Language-Specific Features}

To better understand the emergence and propagation of language-specific representations, we conducted two complementary analyses on the top-ranked SAE features identified across transformer layers: (1) attention head attribution and (2) residual stream decomposition. Both analyses were performed using model inputs comprising 50 sentences sampled from Tatoeba Project (via ManyThings.org) per language, ensuring consistent and controlled comparisons across linguistic settings \cite{tatoeba, manythings}.

First, we analyzed whether specific attention heads contribute disproportionately to language-indicative SAE features within a given layer. For each selected SAE feature, we extracted the output of each attention head before residual addition and computed the dot product between this output and the SAE feature direction. This quantified each head’s alignment with the language feature, allowing us to identify language selective heads that amplify steerable directions in the residual stream.

Second, to trace the origin of language-specific features across layers, we decomposed the residual stream at a given layer into embedding and attention and MLP outputs from prior layers. We then computed the dot product between each component and the current layer’s SAE feature direction to assess how much each upstream block contributed to the construction of the feature. This method followed the residual stream decomposition framework introduced by Elhage et al \cite{elhage2021mathematical}, enabling us to determine whether the steerable direction was introduced locally or inherited from earlier computations. 

Together, these analyses offer a mechanistic understanding of how specific architectural components, both local and upstream, support the emergence and steerability of language-specific representations.

\section{Results}

We evaluate the effectiveness of steering SAE features associated with specific languages across layers of Gemma-2-2B and Gemma-2-9B. Our goal is to assess both the semantic fidelity and linguistic shift induced by these interventions. The results are organized into subsections to highlight the layerwise impact of steering and structural insights from attention heads. For samples of steering results, refer to Table~\ref{tab:steering_examples} and Appendix~\ref{sec:more_steer_example}.

We focus this section on Gemma-2-9B due to its deeper architecture and stronger performance; we present corresponding Gemma-2-2B results in Appendix~\ref{2b_last} and Appendix~\ref{2b_mean}, which show qualitatively similar layerwise trends.

\begin{table*}[t] 
\vskip 0.15in 
\begin{center}
\begin{small} 

\setlength{\tabcolsep}{4pt} 

\begin{tabular}{lcccc} 
\hline
\textbf{Method} & \textbf{CMN} & \textbf{JPN} & \textbf{SPA} & \textbf{FRA} \\
\hline
\multicolumn{5}{l}{\emph{LaBSE Similarity}} \\ 
Steering (Top Feature) & 0.509 $\pm$ 0.016 & 0.435 $\pm$ 0.018 & 0.458 $\pm$ 0.020 & 0.395 $\pm$ 0.021 \\
Prompt Baseline        & 0.277 $\pm$ 0.033 & 0.444 $\pm$ 0.033 & 0.595 $\pm$ 0.029 & 0.477 $\pm$ 0.033 \\
\hline
\multicolumn{5}{l}{\emph{FastText Accuracy}} \\ 
Steering (Top Feature) & 0.978 & 0.938 & 0.888 & 0.852 \\
Prompt Baseline        & 0.356  & 0.598  & 0.786  & 0.626   \\
\hline
\end{tabular}
\caption{Comparison of Sparse Feature Steering and Prompt-Based Baseline on Language Control. 
We report LaBSE similarity scores (95\% confidence interval) and FastText classification accuracy for Chinese (CMN), Japanese (JPN), Spanish (SPA), and French (FRA).}
\label{tab:baseline_comparison}
\end{small}
\end{center}
\vskip -0.1in 
\end{table*}

\subsection{Layerwise Effectiveness of Language Steering}

We evaluate steering effectiveness across layers using two metrics: LaBSE semantic similarity and FastText language classification. For each layer, we apply sparse interventions using the top 3 language-divergent SAE features (selected by last-token activation difference; see Section~\ref{subsec:identify-language-features}) and report the highest-scoring result among them. We focus here on last-token-based feature selection due to its consistently stronger performance across all layers; mean-based results, which exhibit similar trends but lower absolute scores, are shown in Appendix~\ref{9b_mean}.

Figure~\ref{fig:9b_last_sims} presents LaBSE cosine similarity scores between the original English prompt and the steered output. Lines show mean similarity across 500 prompts. Confidence intervals omitted for visual clarity. High scores indicate that steering successfully shifted the model toward the target language while preserving the input’s semantic content. We observe that steering in layers 29–36 yields the strongest preservation for Spanish and French, while Chinese and Japanese show more variability but benefit from similar depths. Earlier layers, by contrast, show minimal effect, confirming that language-relevant features become more concentrated in deeper layers.

Figure~\ref{fig:9b_last_probs} shows the corresponding FastText classification probabilities for the same interventions. Here, high probabilities indicate that the output was identified as being written in the target language. These results largely align with the LaBSE scores: layers with strong semantic preservation also tend to produce outputs recognized as belonging to the intended language. 

To evaluate how sparse feature steering compares with conventional prompting, we introduce a baseline using the instruction-tuned Gemma-2-9B-IT model, which is necessary to ensure the model properly responds to explicit language directives. Each prompt is prepended with “Please generate in [Language]” and 500 English prompts from the Dolly-15k dataset are evaluated for each language. As shown in Table~\ref{tab:baseline_comparison}, steering generally achieves higher target-language classification accuracy, particularly for Chinese (97.8\% vs. 36\%) and Japanese (93.8\% vs. 65\%). While the prompt baseline yields higher LaBSE similarity for all languages except Chinese, steering's performance remain competitive considering its accuracy and additional tuning/prompting is required for comparable performance.

To assess whether sparse feature steering degrades semantic content, we compare LaBSE similarity between two independently generated English outputs from the same prompt (no steering), yielding a baseline of $0.456 \pm 0.019$—reflecting natural variation from decoding stochasticity. When steering into other languages, we observe comparable or even higher semantic similarity between the original English prompt and the steered outputs: Chinese ($0.509 \pm 0.016$), Japanese ($0.435 \pm 0.018$), Spanish ($0.458 \pm 0.020$), and French ($0.395 \pm 0.021$). These results suggest that sparse interventions can preserve semantic meaning across languages, often to a degree similar to or exceeding repeated generation within the same language.

\begin{figure*}[t]
    \centering
    \includegraphics[width=0.8\linewidth]{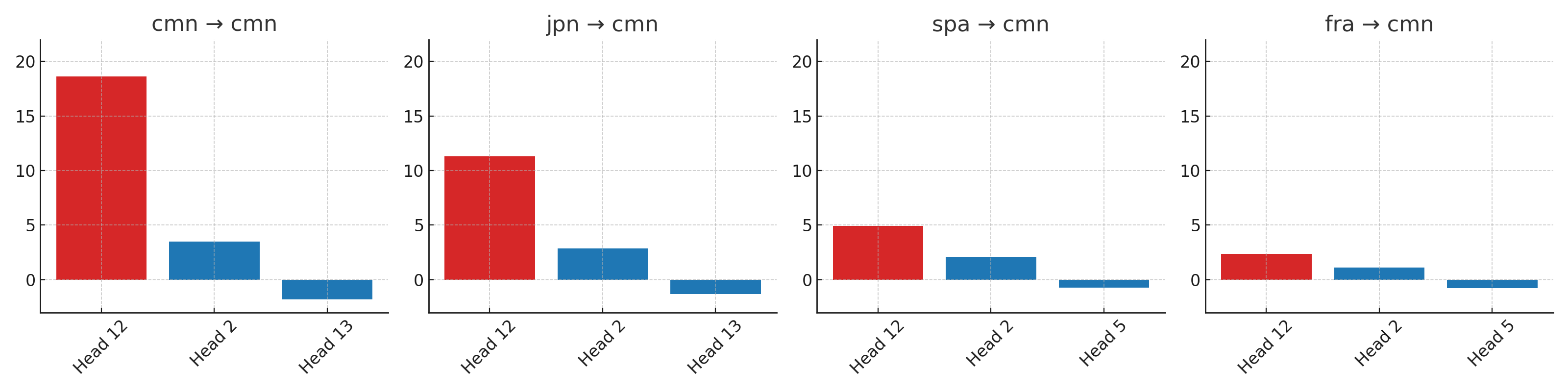}
    \includegraphics[width=0.8\linewidth]{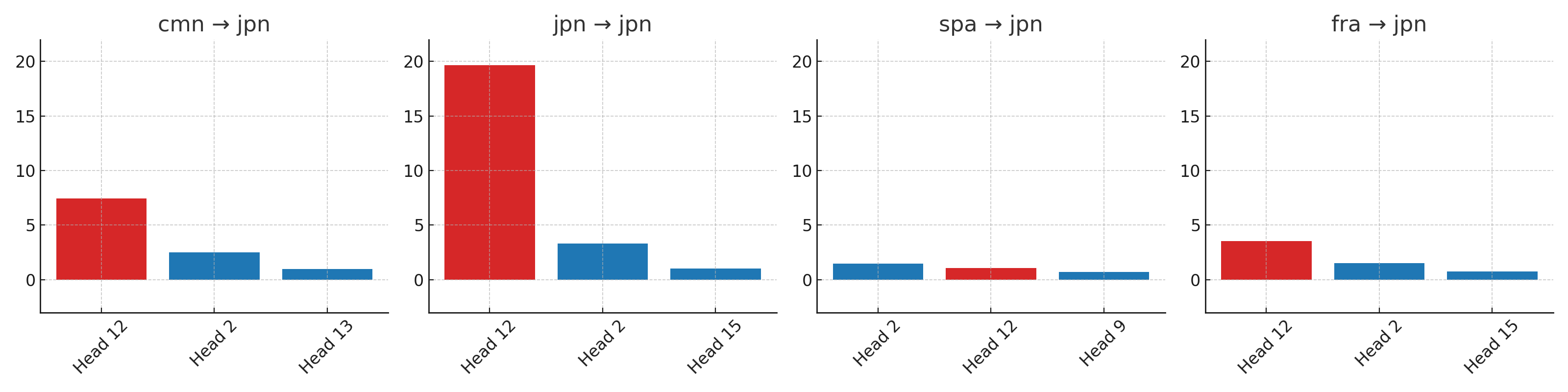}
    \includegraphics[width=0.8\linewidth]{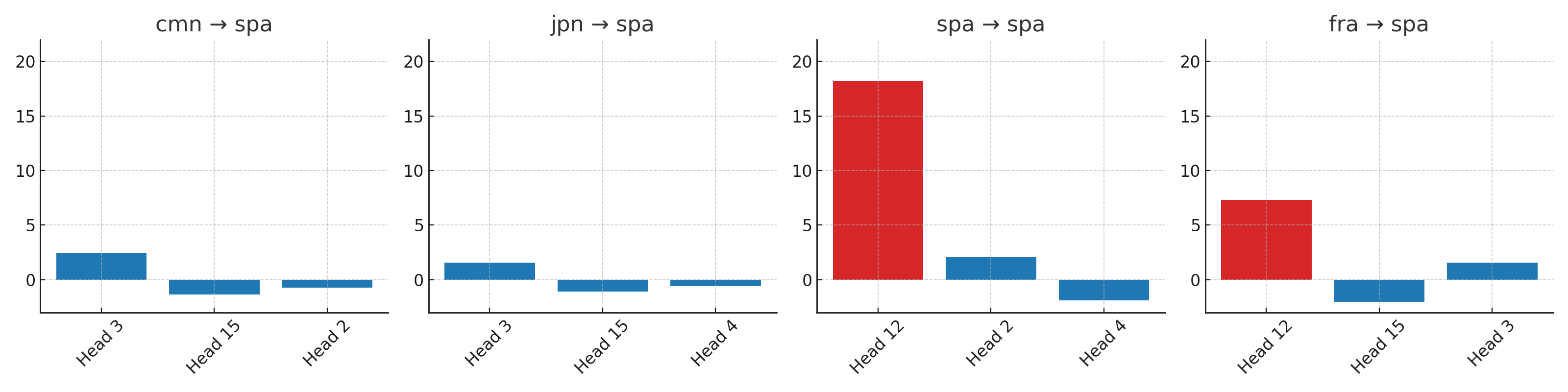}
    \includegraphics[width=0.8\linewidth]{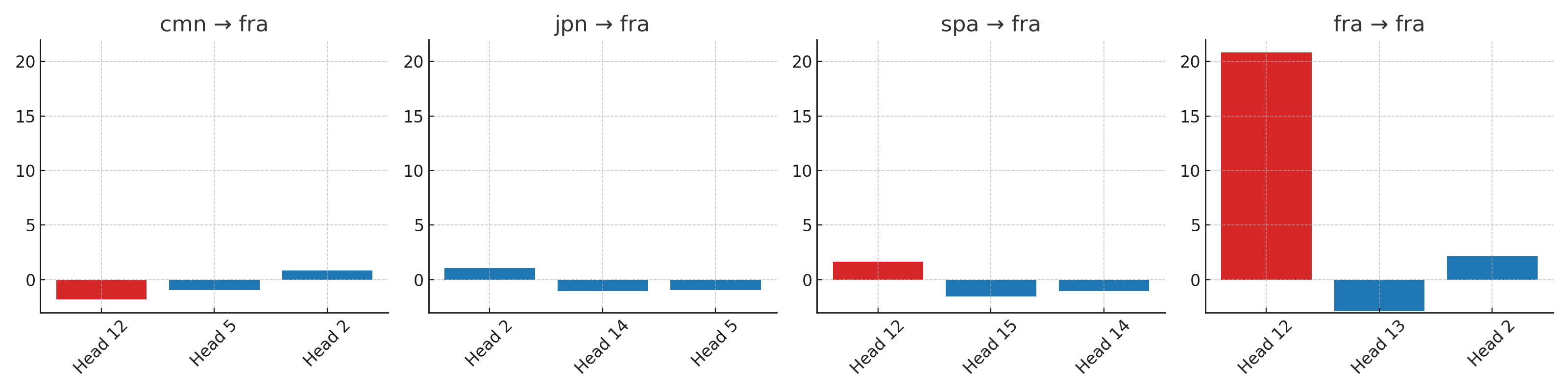}
    \caption{\textbf{Top 3 contributing attention heads at Layer 29 across all Input $\rightarrow$ Feature language pairs.}
    Each subplot shows the three attention heads with the highest contribution to the language-specific SAE feature when the model is given input in a different language. Head 12 is highlighted in red when it appears. The strong, selective dominance of Head 12 in all on-diagonal cases (e.g., \texttt{cmn → cmn}, \texttt{jpn → jpn}, \texttt{fra → fra}) but not off-diagonal cases suggests it plays a role in language-specific representation rather than general-purpose amplification.}
    \label{fig:top-heads-layer-29}
\end{figure*}

\begin{figure*}[t]
    \centering
    \begin{minipage}[t]{0.48\linewidth}
        \centering
        \includegraphics[width=\linewidth]{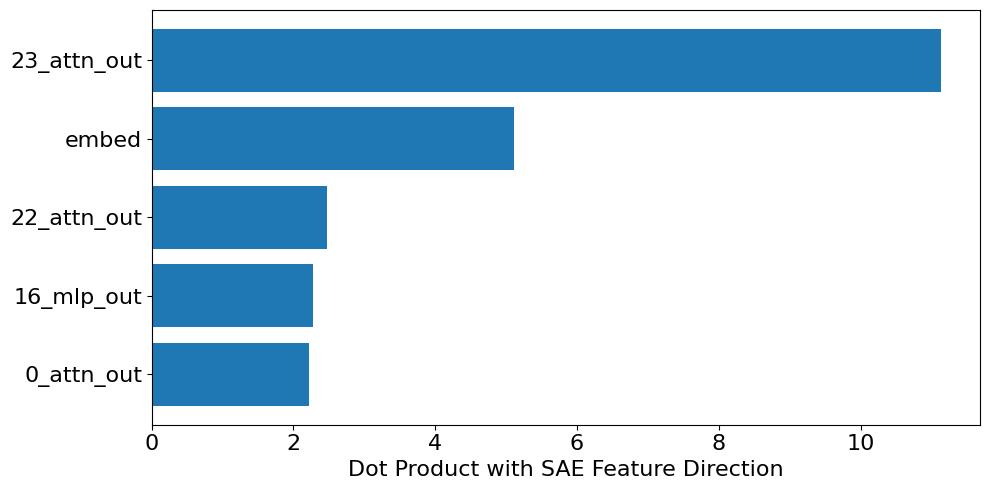}
        \subcaption{Layer 23}
        \label{fig:chinese-feature-contribs-23}
    \end{minipage}
    \hfill
    \begin{minipage}[t]{0.48\linewidth}
        \centering
        \includegraphics[width=\linewidth]{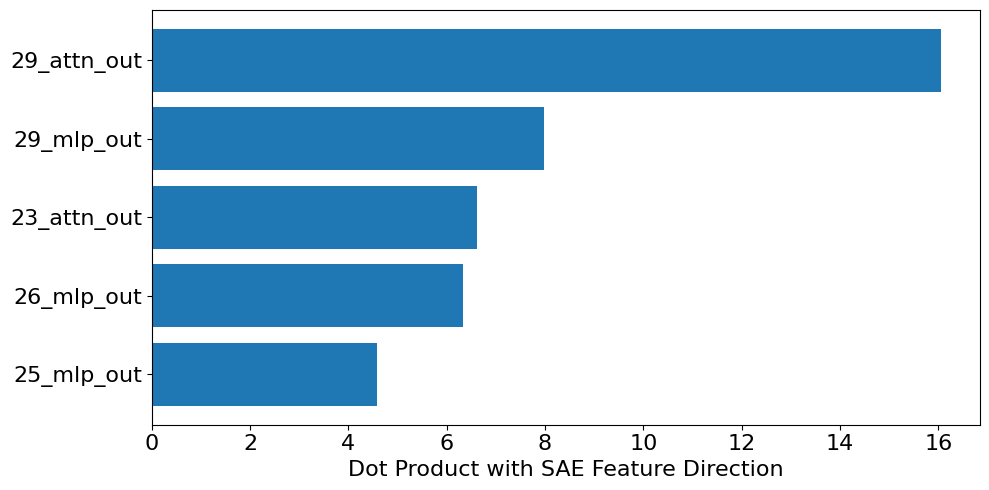}
        \subcaption{Layer 29}
        \label{fig:chinese-feature-contribs-29}
    \end{minipage}
    \caption{
    \textbf{Decomposition of the residual stream in Gemma-2-9B showing the top contributors to the Chinese SAE feature direction at Layers 23 and 29.} Each bar represents the dot product between a residual component (e.g., MLP or attention output from a previous layer) and the language feature direction identified at the current layer. At Layer 23, attention heads in the same layer dominate, suggesting localized construction. At Layer 29, strong contributions also come from within the same layer, but with notable support from Layer 23 and earlier MLPs.
    }
    \label{fig:chinese-feature-contribs-combined}
\end{figure*}

\subsection{Emergence of Language-Specific Attention Heads in 9B}

To better understand the architectural origin of steerable features, we analyzed the attention heads of layers in Gemma-2-9B that exhibited a notable rise in LaBSE semantic similarity, such as layer 29, and measured their contribution to language-representative SAE features at the same layer. Figure~\ref{fig:top-heads-layer-29} reports the top contributing heads for each language at Layer 29 given the input of different languages. 

A consistent pattern emerges: Attention Head 12 contributes significantly more than any other head across all four languages, with dot product between activations and language features reaching 18.61 for Chinese input/feature pair and 20.82 for French input/feature pair. This suggests that Head 12 plays a central role in constructing or amplifying language-aligned representations at this layer. Although steering operates directly on sparse SAE features, our findings indicate that these features may reflect underlying attention-based mechanisms that enhance language specificity in the residual stream.

To further investigate this pattern, we conducted the same analysis at Layers 23 and 30 (see Appendix~\ref{sec:appendix-attn-heads} for supplemental figures). Layer 23 was selected because it also exhibited a marked gain in LaBSE similarity relative to Layer 22. There, we again observed a dominant head, Head 1, contributing strongly to language-specific features, reinforcing the idea that certain layers develop attention-based amplification mechanisms for linguistic identity. In contrast, Layer 30 was selected because, despite its relatively strong steering performance, it did not exhibit a sharp increase in LaBSE similarity from Layer 29. This motivated an investigation into whether its steerability stemmed from inherited features rather than locally concentrated attention contributions.  As seen in Appendix~\ref{top-heads-30}, contributions at Layer 30 were smaller and more diffuse, with no consistent pattern across languages. This suggests that steerability at this depth may arise from distributed representations or inherited features from the previous layer, particularly Layer 29.

Together, these results suggest that while some layers develop explicit mechanisms for steering via language-selective attention heads, others achieve steerability through more diffuse architectural dynamics such as inheritance of feature from previous layers.

\subsection{Tracing the Origin of Language Features Across Layers}

To determine where language-specific SAE features are introduced and amplified, we decomposed the residual stream at layers 23, 29, and 30 into their component sources—attention and MLP outputs from previous layers. We then computed the dot product between each component and the linguistic SAE feature direction at the target layer. We showcase only the Chinese SAE feature as a representative case. As seen in Appendix~\ref{sec:appendix-resid-decomp}, extension to other languages yielded qualitatively similar conclusions.

Our findings reveal that the major increases in steering effectiveness around layers 23 and 29 (as seen in Figure~\ref{fig:9b_last_sims}) align with sharp rises in attention contribution to the language direction, demonstrated in Figure~\ref{fig:chinese-feature-contribs-combined}. In Layer 23, \texttt{23\_attn\_out} dominates, suggesting early insertion of language signal. In Layer 29, \texttt{29\_attn\_out} again leads, confirming that certain attention heads play an amplifying role. However, in layers 30 through 33, the dominant contributors to the residual stream originate from the output of layer 29, which corresponds with the plateau and subsequent decline in semantic steering performance (shown in Figure~\ref{sec:appendix-decomp-30-33}). These results suggest that certain components (attention heads/MLP neurons) at specific layers are the primary source of steerable linguistic representations.

\section{Conclusion}

This work demonstrates that sparse autoencoder (SAE) features in multilingual language models can be causally steered to induce controlled shifts in output language, while preserving semantic intent. By systematically identifying and intervening on language-specific SAE features, we show that steering performance varies across layers and languages, revealing layer-localized sensitivity to linguistic structure. Our analysis further uncovers that certain transformer components in certain high-performing layers—such as Heads 1 and 12 in Layers 23 and 29 of Gemma-2-9B—act as amplifiers of language-correlated directions, while other layers (e.g., Layer 30) achieve similar steering outcomes via more diffuse, non-head-localized mechanisms.

These findings suggest that controlled steering in LLMs engages multiple architectural mechanisms, notably sparse feature activations and attention head specialization. Future directions include generalizing this framework to non-language attributes (e.g., tone, dialect) and extending the interpretability analysis to MLP contributions and other model families.

\section*{Limitations}
\label{sec: limit}

Our study focuses on the analysis of sparse autoencoder (SAE) features in multilingual transformer models and their causal role in steering language generation. While our findings demonstrate consistent trends in the Gemma-2-9B model and are partially supported by Gemma-2-2B, several limitations should be noted:

First, all experiments are conducted exclusively on the Gemma model family, which, while representative of modern multilingual decoders, may limit generalizability to other architectures (e.g., encoder-decoder models or monolingual LLMs) or training paradigms. Further replication across diverse model families would be necessary to establish broader applicability.

Second, the identification of language-divergent SAE features relies on a fixed dataset of 1,000 parallel sentences per language. Steering is evaluated on a curated set of 500 English prompts. Architectural origin of language features is located with 50 sentences per language. While this setup ensures controlled comparisons, it may not capture the full diversity of naturalistic or noisy multilingual input. Our evaluation also depends on automatic metrics—FastText classification and LaBSE similarity—which, although informative, do not fully reflect human judgments of fluency or meaning preservation.

Third, the SAE features used for steering are derived from a pretrained sparse autoencoder and are not learned end-to-end for the steering task. This limits their optimality and may affect performance bounds. Additionally, while our decomposition and attention attribution analyses offer mechanistic insights, they remain correlational and rely on assumptions about linear contributions to the residual stream. The directionality of influence (e.g., whether attention heads induce feature alignment or respond to it) is not conclusively established.

Finally, the computational requirements of our method are nontrivial for an inference task. All experiments used a single NVIDIA A100 SXM 80GB GPU and completed within~14 hours per language. Scaling our method to a larger LLM would also require training SAEs across all layers, which is computationally expensive.

We encourage future work to explore the generality of these findings across models, expand the interpretability of SAE features beyond language identity, and investigate training-time interventions or more efficient steering mechanisms.

\section*{Acknowledgement}
We thank Rithvik Chigurupati and Ronoy Sarkar for their helpful feedback throughout the work.

\bibliography{custom}

\begin{thebibliography}{15}
\providecommand{\natexlab}[1]{#1}

\bibitem[{Bricken et~al.(2023)Bricken, Templeton, Batson, Chen, Jermyn, Conerly, Turner, Anil, Denison, Askell, Lasenby, Wu, Kravec, Schiefer, Maxwell, Joseph, Hatfield-Dodds, Tamkin, Nguyen, McLean, Burke, Hume, Carter, Henighan, and Olah}]{bricken2023monosemanticity}
Trenton Bricken, Adly Templeton, Joshua Batson, Brian Chen, Adam Jermyn, Tom Conerly, Nick Turner, Cem Anil, Carson Denison, Amanda Askell, Robert Lasenby, Yifan Wu, Shauna Kravec, Nicholas Schiefer, Tim Maxwell, Nicholas Joseph, Zac Hatfield-Dodds, Alex Tamkin, Karina Nguyen, and 6 others. 2023.
\newblock Towards monosemanticity: Decomposing language models with dictionary learning.
\newblock \emph{Transformer Circuits Thread}.
\newblock Https://transformer-circuits.pub/2023/monosemantic-features/index.html.

\bibitem[{Chalnev et~al.(2024)Chalnev, Siu, and Conmy}]{chalnev2024saets}
Sergey Chalnev, Michael Siu, and Alexander Conmy. 2024.
\newblock \href {https://doi.org/10.48550/arXiv.2411.02193} {Improving steering vectors by targeting sparse autoencoder features}.
\newblock \emph{Preprint}, arXiv:2411.02193.
\newblock ArXiv preprint.

\bibitem[{Chang et~al.(2022)Chang, Tu, and Bergen}]{chang-etal-2022-geometry}
Tyler Chang, Zhuowen Tu, and Benjamin Bergen. 2022.
\newblock \href {https://doi.org/10.18653/v1/2022.emnlp-main.9} {The geometry of multilingual language model representations}.
\newblock In \emph{Proceedings of the 2022 Conference on Empirical Methods in Natural Language Processing}, pages 119--136, Abu Dhabi, United Arab Emirates. Association for Computational Linguistics.

\bibitem[{Charles and Lawrence(2024)}]{manythings}
Kelly Charles and Kelly Lawrence. 2024.
\newblock Manythings.org: English sentence pairs from the tatoeba project.
\newblock \url{https://www.manythings.org/anki/}.
\newblock Accessed: 2025-05-10.

\bibitem[{Conover et~al.(2023)Conover, Hayes, Mathur, Xie, Wan, Shah, Ghodsi, Wendell, Zaharia, and Xin}]{DatabricksBlog2023DollyV2}
Mike Conover, Matt Hayes, Ankit Mathur, Jianwei Xie, Jun Wan, Sam Shah, Ali Ghodsi, Patrick Wendell, Matei Zaharia, and Reynold Xin. 2023.
\newblock \href {https://www.databricks.com/blog/2023/04/12/dolly-first-open-commercially-viable-instruction-tuned-llm} {Free dolly: Introducing the world's first truly open instruction-tuned llm}.

\bibitem[{Elhage et~al.(2022)Elhage, Hume, Olsson, Schiefer, Henighan, Kravec, Hatfield-Dodds, Lasenby, Drain, Chen, Grosse, McCandlish, Kaplan, Amodei, Wattenberg, and Olah}]{elhage2022superposition}
Nelson Elhage, Tristan Hume, Catherine Olsson, Nicholas Schiefer, Tom Henighan, Shauna Kravec, Zac Hatfield-Dodds, Robert Lasenby, Dawn Drain, Carol Chen, Roger Grosse, Sam McCandlish, Jared Kaplan, Dario Amodei, Martin Wattenberg, and Christopher Olah. 2022.
\newblock Toy models of superposition.
\newblock \emph{Transformer Circuits Thread}.
\newblock Https://transformer-circuits.pub/2022/toy\_model/index.html.

\bibitem[{Elhage et~al.(2021)Elhage, Nanda, Olsson, Henighan, Joseph, Mann, Askell, Bai, Chen, Conerly, DasSarma, Drain, Ganguli, Hatfield-Dodds, Hernandez, Jones, Kernion, Lovitt, Ndousse, Amodei, Brown, Clark, Kaplan, McCandlish, and Olah}]{elhage2021mathematical}
Nelson Elhage, Neel Nanda, Catherine Olsson, Tom Henighan, Nicholas Joseph, Ben Mann, Amanda Askell, Yuntao Bai, Anna Chen, Tom Conerly, Nova DasSarma, Dawn Drain, Deep Ganguli, Zac Hatfield-Dodds, Danny Hernandez, Andy Jones, Jackson Kernion, Liane Lovitt, Kamal Ndousse, and 6 others. 2021.
\newblock A mathematical framework for transformer circuits.
\newblock \emph{Transformer Circuits Thread}.
\newblock Https://transformer-circuits.pub/2021/framework/index.html.

\bibitem[{Feng et~al.(2022)Feng, Yang, Cer, Arivazhagan, and Wang}]{feng-etal-2022-language}
Fangxiaoyu Feng, Yinfei Yang, Daniel Cer, Naveen Arivazhagan, and Wei Wang. 2022.
\newblock \href {https://doi.org/10.18653/v1/2022.acl-long.62} {Language-agnostic {BERT} sentence embedding}.
\newblock In \emph{Proceedings of the 60th Annual Meeting of the Association for Computational Linguistics (Volume 1: Long Papers)}, pages 878--891, Dublin, Ireland. Association for Computational Linguistics.

\bibitem[{Huben et~al.(2024)Huben, Cunningham, Smith, Ewart, and Sharkey}]{huben24}
Robert Huben, Hoagy Cunningham, Logan~Riggs Smith, Aidan Ewart, and Lee Sharkey. 2024.
\newblock \href {https://openreview.net/forum?id=F76bwRSLeK} {Sparse autoencoders find highly interpretable features in language models}.
\newblock In \emph{The Twelfth International Conference on Learning Representations}.

\bibitem[{Joulin et~al.(2016)Joulin, Grave, Bojanowski, and Mikolov}]{joulin2016bag}
Armand Joulin, Edouard Grave, Piotr Bojanowski, and Tomas Mikolov. 2016.
\newblock Bag of tricks for efficient text classification.
\newblock \emph{arXiv preprint arXiv:1607.01759}.

\bibitem[{Lieberum et~al.(2024)Lieberum, Rajamanoharan, Conmy, Smith, Sonnerat, Varma, Kramár, Dragan, Shah, and Nanda}]{lieberum2024gemmascopeopensparse}
Tom Lieberum, Senthooran Rajamanoharan, Arthur Conmy, Lewis Smith, Nicolas Sonnerat, Vikrant Varma, János Kramár, Anca Dragan, Rohin Shah, and Neel Nanda. 2024.
\newblock \href {https://arxiv.org/abs/2408.05147} {Gemma scope: Open sparse autoencoders everywhere all at once on gemma 2}.
\newblock \emph{Preprint}, arXiv:2408.05147.

\bibitem[{Soo et~al.(2025)Soo, Teng, Balaganesh, Guoxian, and YAN}]{soo2025interpretable}
Samuel Soo, Wesley Teng, Chandrasekaran Balaganesh, Tan Guoxian, and Ming YAN. 2025.
\newblock \href {https://openreview.net/forum?id=swRxS7s4rB} {Interpretable steering of large language models with feature guided activation additions}.
\newblock In \emph{ICLR 2025 Workshop on Building Trust in Language Models and Applications}.

\bibitem[{Tang et~al.(2024)Tang, Luo, Huang, Zhang, Wang, Zhao, Wei, and Wen}]{tang-etal-2024-language}
Tianyi Tang, Wenyang Luo, Haoyang Huang, Dongdong Zhang, Xiaolei Wang, Xin Zhao, Furu Wei, and Ji-Rong Wen. 2024.
\newblock \href {https://doi.org/10.18653/v1/2024.acl-long.309} {Language-specific neurons: The key to multilingual capabilities in large language models}.
\newblock In \emph{Proceedings of the 62nd Annual Meeting of the Association for Computational Linguistics (Volume 1: Long Papers)}, pages 5701--5715, Bangkok, Thailand. Association for Computational Linguistics.

\bibitem[{Team(2024)}]{gemma_2024}
Gemma Team. 2024.
\newblock \href {https://doi.org/10.34740/KAGGLE/M/3301} {Gemma}.

\bibitem[{{The Tatoeba Project Contributors}(2024)}]{tatoeba}
{The Tatoeba Project Contributors}. 2024.
\newblock The tatoeba project: Multilingual sentence dataset.
\newblock \url{https://tatoeba.org}.
\newblock Creative Commons Attribution 2.0 France (CC-BY 2.0 FR). Sentence authors listed at: \url{https://tatoeba.org}.

\end{thebibliography}

\appendix
\onecolumn
\section{Additional Examples of Multilingual Language Steering}
\label{sec:more_steer_example}
We show more Gemma-2-9B's continuation for a single prompt when steered on the best SAE feature for each language, scored by method detailed in section~\ref{Evaluation Metrics for Output Quality}. Due to max token constraint of 64, the final unfinished sentence is removed for clarity.
\begin{table}[H]
  \centering
  \begin{tabular}{p{3cm} | p{12cm}}
    \hline
    \textbf{Prompt} & \texttt{A just society should provide} \\
    \hline
    \textbf{No Steer} &
    A just society should provide for the needs of its population, and not just those who are able to earn a living. In our country, the government does this through a social safety net that includes various programs such as Old Age Security, Employment Insurance and the Canada Pension Plan. \\
    \hline
    \textbf{Steered (Chinese)} &
    \begin{CJK}{UTF8}{gbsn}A just society should provide机会给每个人，无论他们来自哪里或他们的背景。然而，在现实生活中，许多人发现自己处于不利地位。我们看到这个在教育系统中。有证据表明，如果一个学生来自一个贫穷家庭，他/她在学校的表现可能会受到影响。\end{CJK} \\
    \hline
    \textbf{Steered (Spanish)} &
    A just society should provide for la educación y la capacitación de todos los ciudadanos en condiciones de igualdad. A medida que crezcamos, las oportunidades para aprender deben seguir siendo accesibles y accesibles. Es por eso que la educación y la capacitación a lo largo de toda la vida son el núcleo de nuestro programa. \\
    \hline
    \textbf{Steered (Japanese)} &
    \begin{CJK}{UTF8}{min}A just society should provide教育で学び、社会のルールを理解する機会を与える必要があります。「人権と社会正義」の授業では、生徒が「社会正義」について考える機会を設けました。授業では、生徒は問題解決法について話し合いました。\end{CJK} \\
    \hline
    \textbf{Steered (French)} &
    A just society should provide for la justice sociale doit assurer la protection et le soutien nécessaires aux personnes vulnérables, notamment les enfants, les femmes et les personnes âgées.La justice sociale doit également promouvoir l’égalité des chances et la participation de tous à la vie économique, sociale et politique de la société. \\
    \hline
  \end{tabular}
\end{table}
\clearpage
\begin{table}[H]
  \centering
  \begin{tabular}{p{3cm} | p{12cm}}
    \hline
    \textbf{Prompt} & \texttt{Online education has changed learning by} \\
    \hline
    \textbf{No Steer} &
    Online education has changed learning by allowing students to access educational material at any time and place. However, the quality of online education is generally lower than that of traditional face-to-face education. The purpose of this study was to investigate the effects of gamification on the quality of online education. \\
    \hline
    \textbf{Steered (\mbox{Chinese})} &
    \begin{CJK}{UTF8}{gbsn}Online education has changed learning by 允许学生在任何地方学习。与传统教育相比，在线学习提供更多灵活性，并允许学生根据自己的时间表上课。它还提供了一个更大的课程选择，允许学生选择他们感兴趣的主题。在线教育也为教师提供了新的教学方法。\end{CJK} \\
    \hline
    \textbf{Steered (\mbox{Spanish})} &
    Online education has changed learning by allowing students to access educational material y donde quiera que estén. En línea, los estudiantes pueden aprender a su propio ritmo, sin tener que asistir a clases presenciales o preocuparse por la asistencia. Esto también significa que los estudiantes pueden aprender de manera más flexible y adaptarse a sus horarios. \\
    \hline
    \textbf{Steered (Japanese)} &
    \begin{CJK}{UTF8}{min}Online education has changed learning by allowing students to learn from anywhere。It は、場所を問わず学習できるようになり、学習方法を変えました。Itは、学生にとってはより便利な方法で勉強できるようになりました。Theyは、授業をオンラインで受けられるようになったので、授業に出なくても勉強できるようになりました。\end{CJK} \\
    \hline
    \textbf{Steered (French)} &
    Online education has changed learning by permettant à tous les étudiants, peu importe leur emplacement géographique, de suivre des cours à distance. Le marché mondial de l’éducation en ligne est en pleine croissance et a été évalué à 191 milliards de dollars en 2020. \\
    \hline
  \end{tabular}
\end{table}
\begin{table}[h]
  \centering
  \begin{tabular}{p{3cm} | p{12cm}}
    \hline
    \textbf{Prompt} & \texttt{Replication in experiments ensures} \\
    \hline
    \textbf{No Steer} &
    Replication in experiments ensures that the same procedure is applied to all samples. This is particularly important in the biological sciences, where there are many variables that can influence an experiment. The same procedure can be applied to different samples, or the same sample can be tested multiple times.  \\
    \hline
    \textbf{Steered (\mbox{Chinese})} &
    \begin{CJK}{UTF8}{gbsn}Replication in experiments ensures A.高重复性 B.高精确度 C.客观性 D.可重复性 在以下哪项中，不属于风电场设备的组成部件？A.风机叶片B.风塔支架 C.变电站 D.电\end{CJK} \\
    \hline
    \textbf{Steered (\mbox{Spanish})} &
    Replication in experiments ensures que la selección de los datos no esté sesgada en favor de un determinado resultado. La replicación es un requisito para la validez científica. Los métodos de investigación cuantitativa y cualitativa pueden ser replicables.  \\
    \hline
    \textbf{Steered (Japanese)} &
    \begin{CJK}{UTF8}{min}Replication in experiments ensures(a)再現実験の目的は実験を繰り返すことである(b)再現実験は、結果を正当なものにするものである。(c)再現実験は、科学的検証の一環である。(d)再現実験は、科学的検証の一環である。（正解）\end{CJK} \\
    \hline
    \textbf{Steered (French)} &
    Replication in experiments ensures que la science est une entreprise humaine. Les scientifiques se sont efforcés de s’assurer que les résultats de leurs recherches soient reproductibles. De nombreuses études ont été reproduites avec des résultats similaires. \\
    \hline
  \end{tabular}
\end{table}
\begin{table}[H]
  \centering
  \begin{tabular}{p{3cm} | p{12cm}}
    \hline
    \textbf{Prompt} & \texttt{Inflation affects the society by} \\
    \hline
    \textbf{No Steer} &
    Inflation affects the society by affecting the price of goods, services, and money. The prices of goods and services rise as inflation rises. As a result, purchasing power decreases. The value of money is reduced as well. <strong>References</strong> https://www.investopedia.com/terms/i/inflation.asp \\
    \hline
    \textbf{Steered (\mbox{Chinese})} &
    \begin{CJK}{UTF8}{gbsn}Inflation affects the society by导致社会对高通货膨胀的反应。人们有不同的反应，但最常见的反应是通货膨胀会导致人们失去信任政府。由于通风管道，人们将他们对政府的信任从一个政府转移到另一个政府。这导致了政治不稳定，并最终导致政变。\end{CJK} \\
    \hline
    \textbf{Steered (\mbox{Spanish})} &
    Inflation affects the society by:
* Destrucción de la riqueza.
* Disminución del poder adquisitivo.
* Desaparición de la inversión.
La inflación es el aumento generalizado y sostenido de los precios de los bienes y servicios en una economía durante un periodo determinado. \\
    \hline
    \textbf{Steered (Japanese)} &
    \begin{CJK}{UTF8}{min}Inflation affects the society by影響社会の【影響】「影響」は「影響を受ける」で意味と発音が分かります。「影響する」は「～に影響する」「～に悪い影響を与える」「～に良い影響を与える」など様々なパターンがあります。【社会】\end{CJK} \\
    \hline
    \textbf{Steered (French)} &
    Inflation affects the society by affecting la production et la consommation, les échanges commerciaux et les taux de change. La production est l'ensemble des biens et services que l'on met sur le marché. Les ménages produisent ces biens à partir de ressources naturelles, de travail et de capital. La société produit pour répondre aux besoins des consommateurs.  \\
    \hline
  \end{tabular}
\end{table}
\FloatBarrier

\section{Layerwise Steering Performance}

\subsection{Mean-Activation-Selected Features from Gemma-2-9B}
\label{9b_mean}
\FloatBarrier
\begin{figure}[H]
  \centering
  \begin{minipage}[t]{0.49\linewidth}
    \centering
    \includegraphics[width=\linewidth]{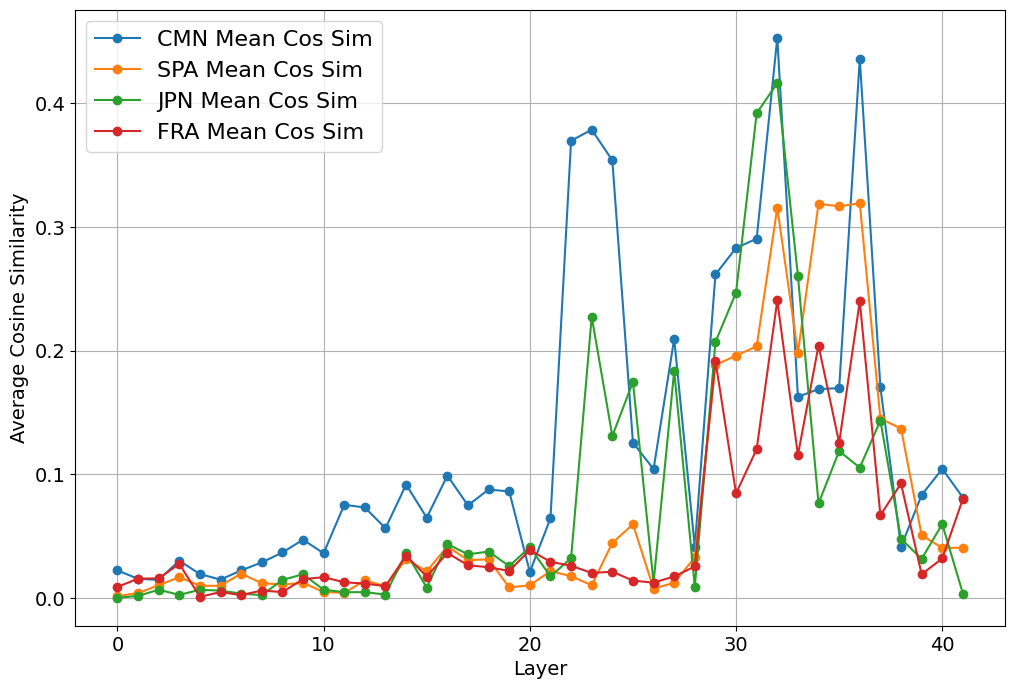}
    \caption{LaBSE semantic similarity scores for steered outputs across Gemma-2-9B layers, using mean-activation-selected language features for Chinese (CMN), Spanish (SPA), Japanese (JPN), and French (FRA). The results show that steering effectiveness varies across layers, with peak semantic alignment occurring at mid to late layers for different languages.}
    \label{fig:9b_mean_sims}
  \end{minipage}%
  \hfill
  \begin{minipage}[t]{0.49\linewidth}
    \centering
    \includegraphics[width=\linewidth]{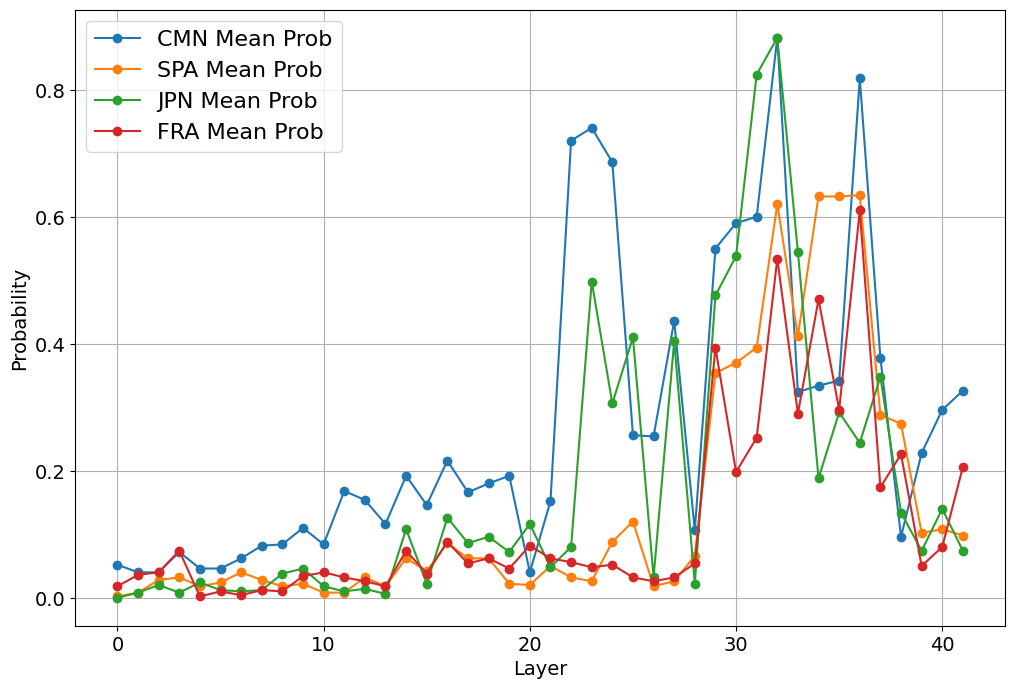}
    \caption{FastText classification probabilities of the same steered outputs, revealing layer-specific differences in how strongly outputs reflect the target language. Later layers generally show higher classification confidence, indicating greater controllability through steering at those depths.}
    \label{fig:9b_mean_probs}
  \end{minipage}
\end{figure}
\FloatBarrier

\subsection{Last-Token-Selected Features from Gemma-2-2B}
\label{2b_last}
\begin{figure}[H]
  \centering
  \begin{minipage}[t]{0.49\linewidth}
    \centering
    \includegraphics[width=\linewidth]{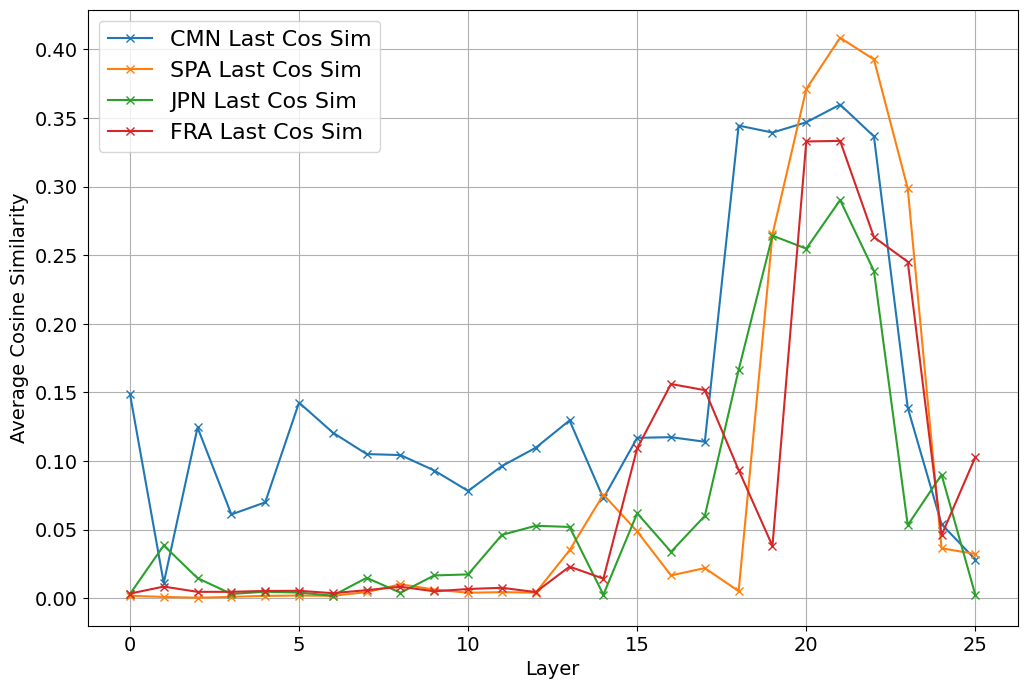}
    \caption{LaBSE semantic similarity scores for steered outputs across Gemma-2-2B layers, using last-token-selected language features for Chinese (CMN), Spanish (SPA), Japanese (JPN), and French (FRA). The results show that steering effectiveness varies across layers, with peak semantic alignment occurring at mid to late layers for different languages.}
    \label{fig:2b_last_sims}
  \end{minipage}%
  \hfill
  \begin{minipage}[t]{0.49\linewidth}
    \centering
    \includegraphics[width=\linewidth]{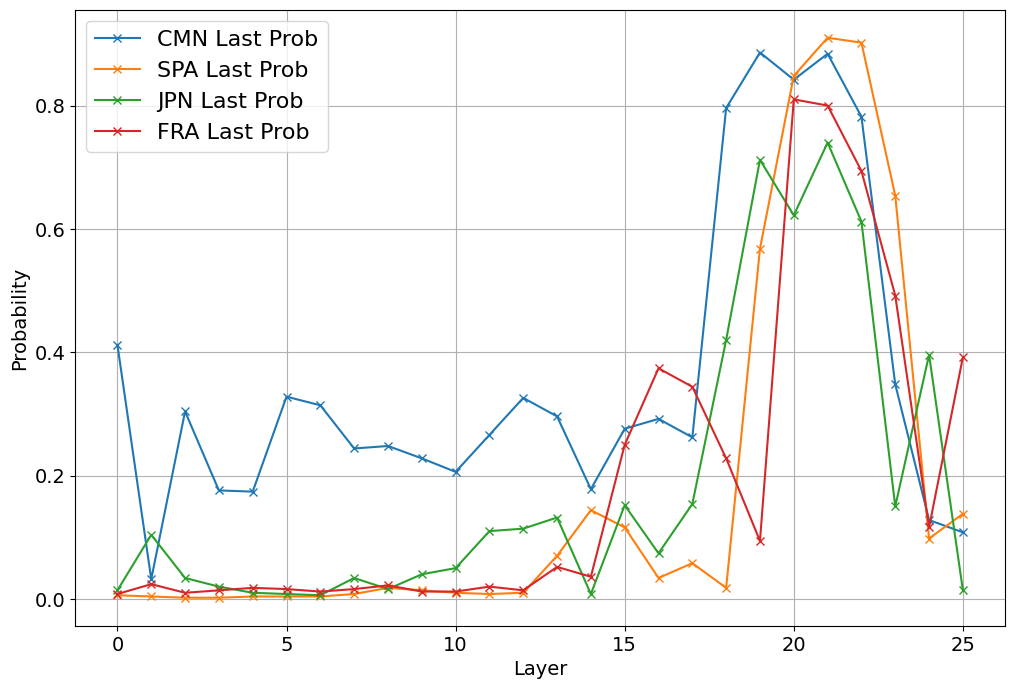}
    \caption{FastText classification probabilities of the same steered outputs, revealing layer-specific differences in how strongly outputs reflect the target language. Later layers generally show higher classification confidence, indicating greater controllability through steering at those depths.}
    \label{fig:2b_last_probs}
  \end{minipage}
\end{figure}
\FloatBarrier

\subsection{Mean-Activation-Selected Features from Gemma-2-2B}
\label{2b_mean}
\begin{figure}[H]
  \centering
  \begin{minipage}[t]{0.49\linewidth}
    \centering
    \includegraphics[width=\linewidth]{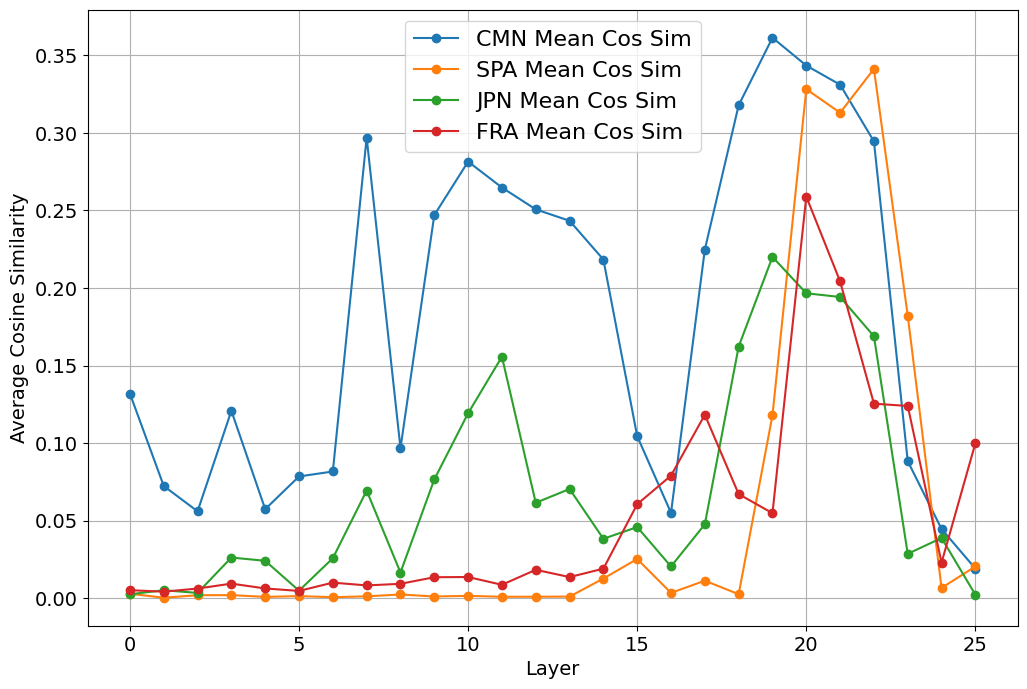}
    \caption{LaBSE semantic similarity scores for steered outputs across Gemma-2-2B layers, using mean-activation-selected language features for Chinese (CMN), Spanish (SPA), Japanese (JPN), and French (FRA). The results show that steering effectiveness varies across layers, with peak semantic alignment occurring at mid to late layers for different languages.}
    \label{fig:2b_mean_sims}
  \end{minipage}%
  \hfill
  \begin{minipage}[t]{0.49\linewidth}
    \centering
    \includegraphics[width=\linewidth]{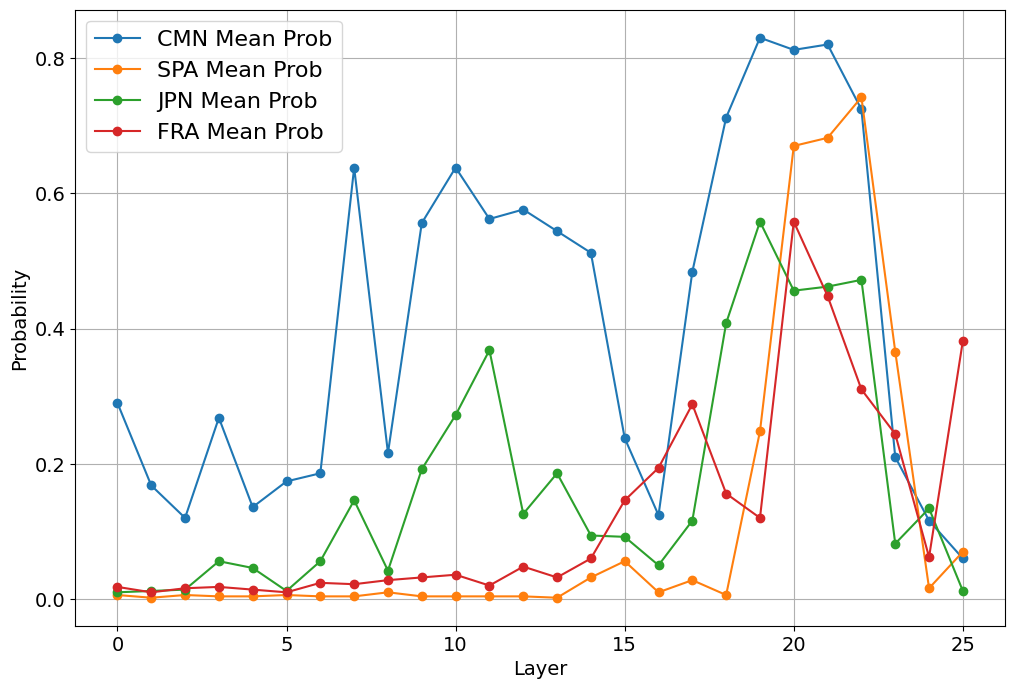}
    \caption{FastText classification probabilities of the same steered outputs, revealing layer-specific differences in how strongly outputs reflect the target language. Later layers generally show higher classification confidence, indicating greater controllability through steering at those depths.}
    \label{fig:2b_mean_probs}
  \end{minipage}
\end{figure}
\FloatBarrier

\section{Language-Specific Attention Contributions at Layers 23 and 30}
\label{sec:appendix-attn-heads}
\subsection{Layer 23}
\label{top-heads-23}
\begin{figure}[H]
    \centering
    \includegraphics[width=\linewidth]{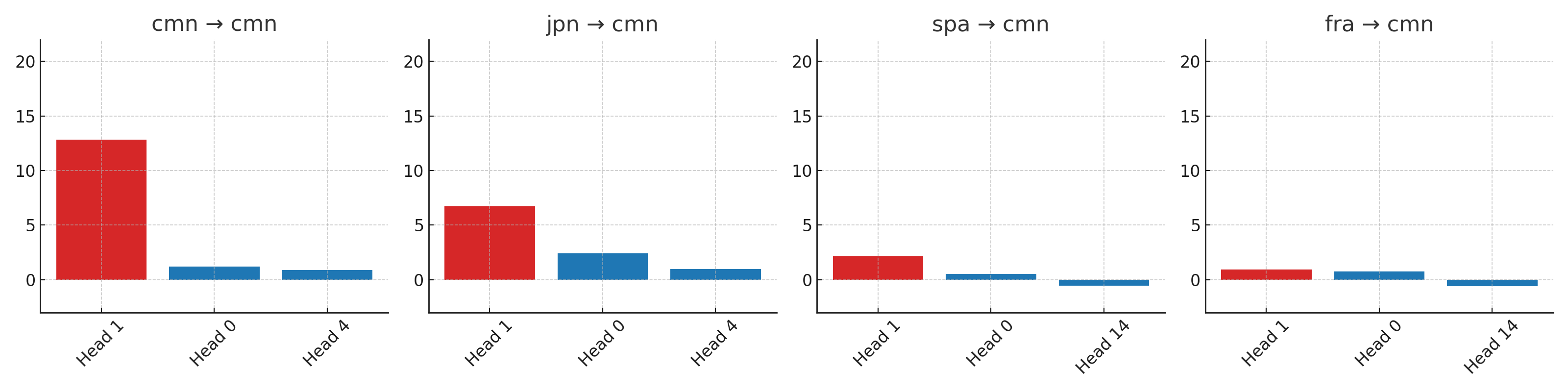}
    \includegraphics[width=\linewidth]{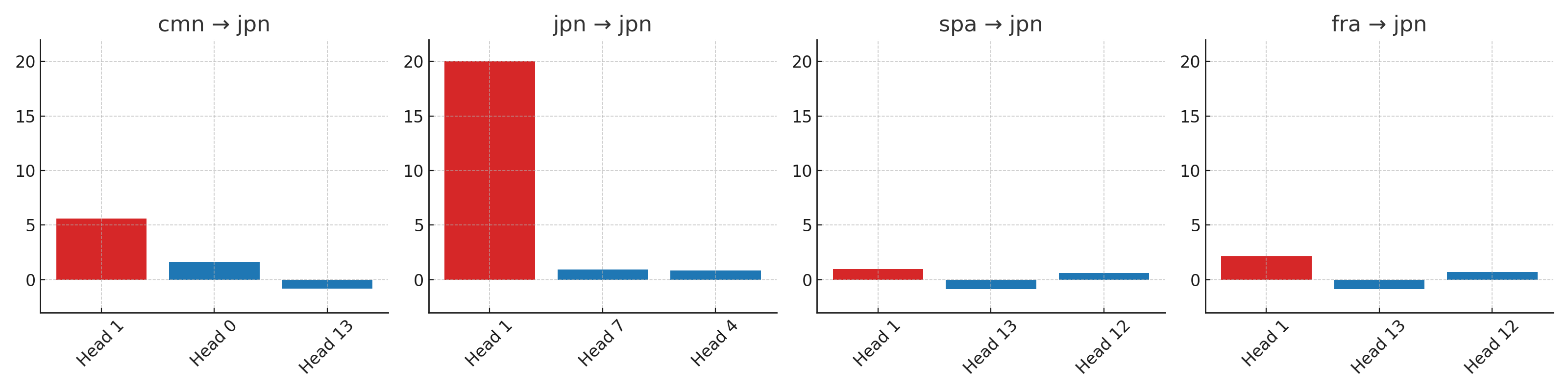}
    \includegraphics[width=\linewidth]{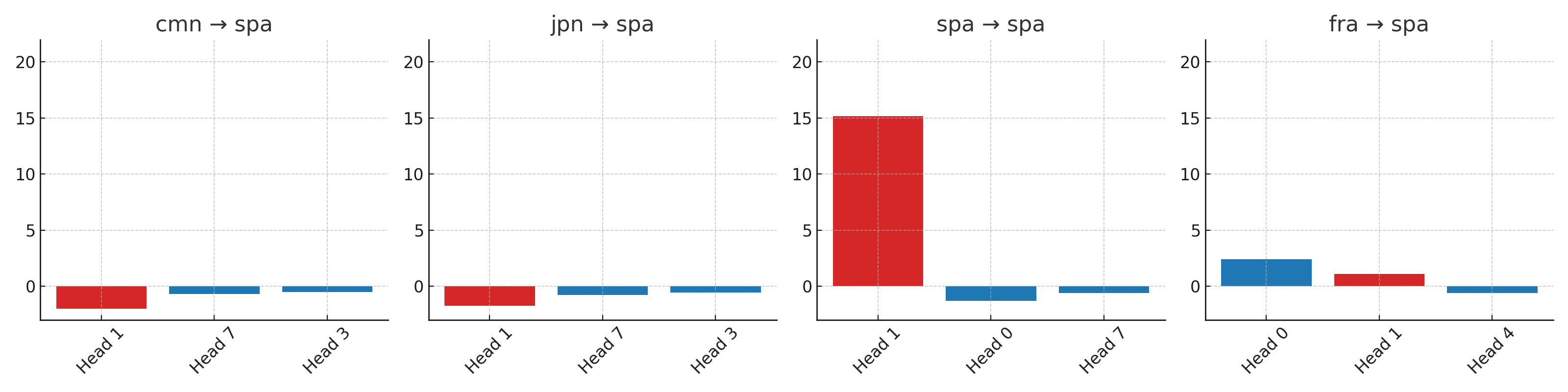}
    \includegraphics[width=\linewidth]{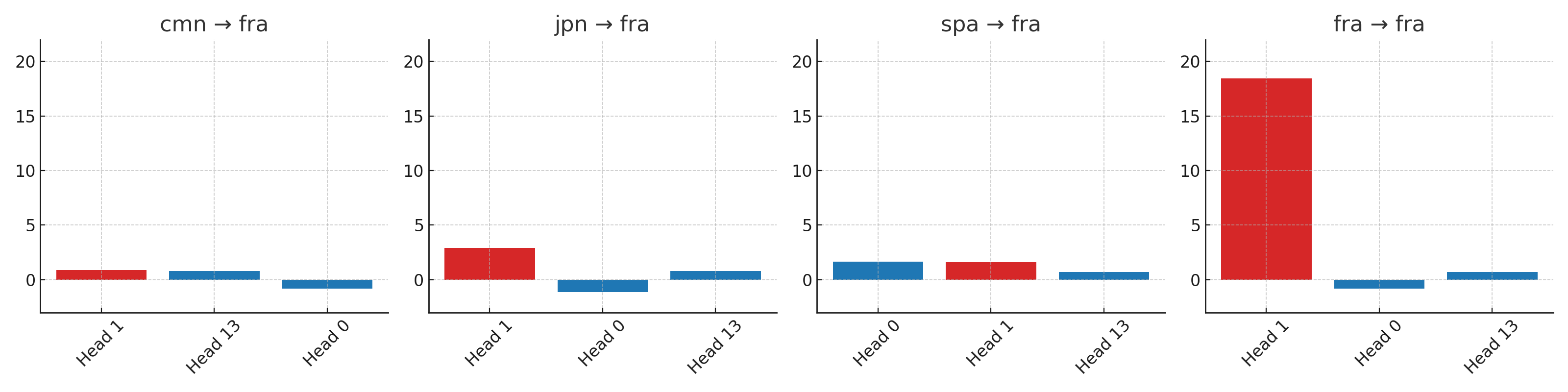}
    \caption{\textbf{Top 3 contributing attention heads at Layer 23 across all Input $\rightarrow$ Feature language pairs.}
    Each subplot shows the three attention heads with the highest contribution to the language-specific SAE feature when the model is given input in a different language. Head 1 is highlighted in red when it appears. The strong, selective dominance of Head 1 in all on-diagonal cases (e.g., \texttt{cmn → cmn}, \texttt{jpn → jpn}, \texttt{fra → fra}) but not off-diagonal cases suggests it plays a role in language-specific representation rather than general-purpose amplification.}
    \label{fig:top-heads-layer-23}
\end{figure}
\FloatBarrier

\subsection{Layer 30}
\label{top-heads-30}
\begin{figure}[H]
    \centering
    \includegraphics[width=\linewidth]{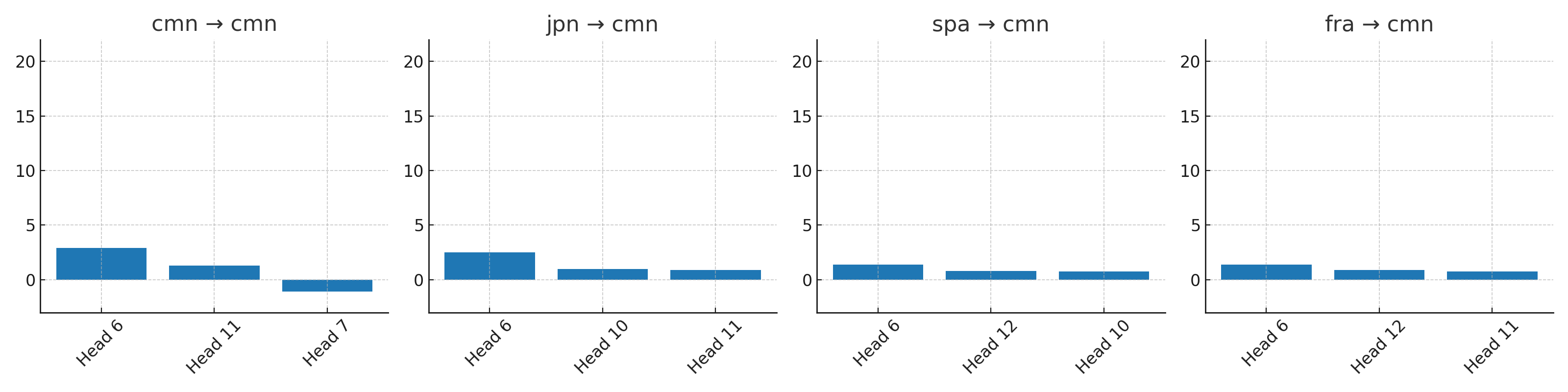}
    \includegraphics[width=\linewidth]{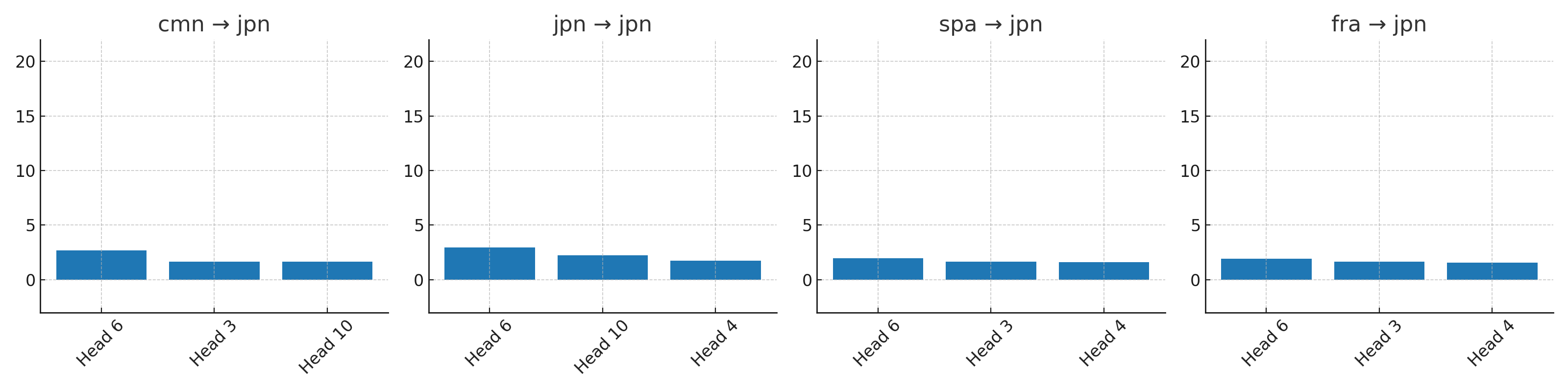}
    \includegraphics[width=\linewidth]{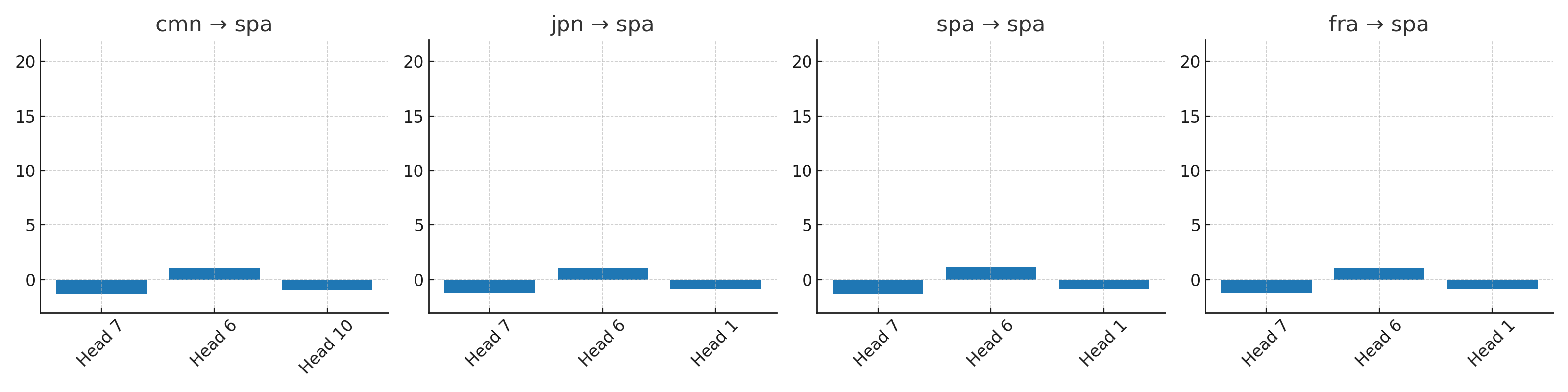}
    \includegraphics[width=\linewidth]{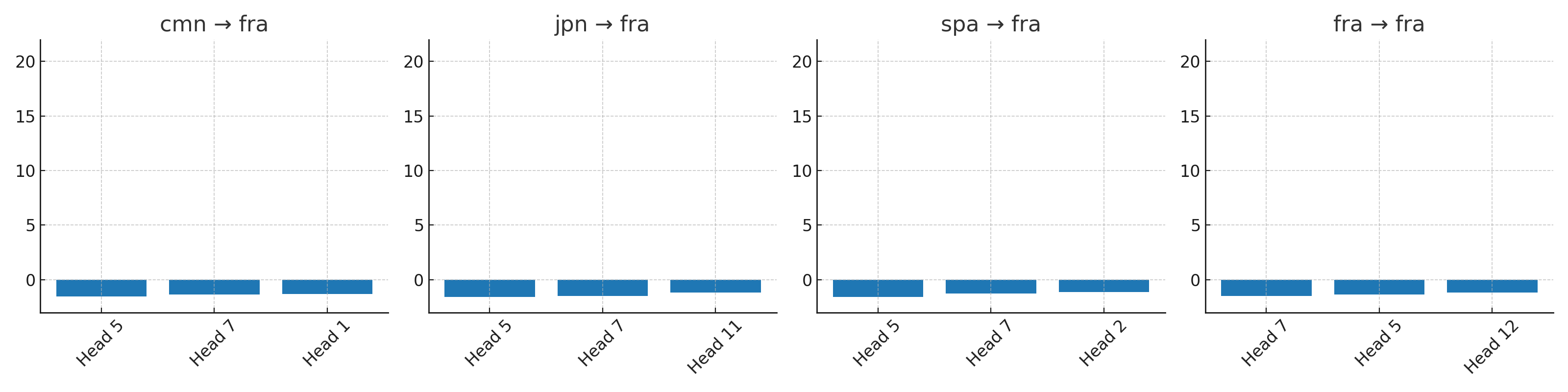}
    \caption{\textbf{Top 3 contributing attention heads at Layer 30 across all Input $\rightarrow$ Feature language pairs.}
    Each subplot shows the three attention heads with the highest contribution to the language-specific SAE feature when the model is given input in a different language. There are no dominance of a certain head, suggesting that no attention heads focus on language-specific amplification and the high steering scores could be inherited from previous layers.}
    \label{fig:top-heads-layer-30}
\end{figure}
\FloatBarrier

\section{Residual Stream Decomposition Analyses}
\label{sec:appendix-resid-decomp}

\subsection{Language Feature Construction in Layers 23 and 29}
\label{sec:appendix-decomp-23-29}

We provide detailed residual stream decomposition results for the Chinese SAE feature at Layers 23 and 29, as shown in Figure~\ref{fig:chinese-feature-contribs-combined} in the main text. Below, we extend this analysis to Spanish, French, and Japanese at the same layers.

\subsubsection{Japanese}
\begin{figure}[H]
    \centering
    \begin{minipage}[t]{0.48\linewidth}
        \centering
        \includegraphics[width=\linewidth]{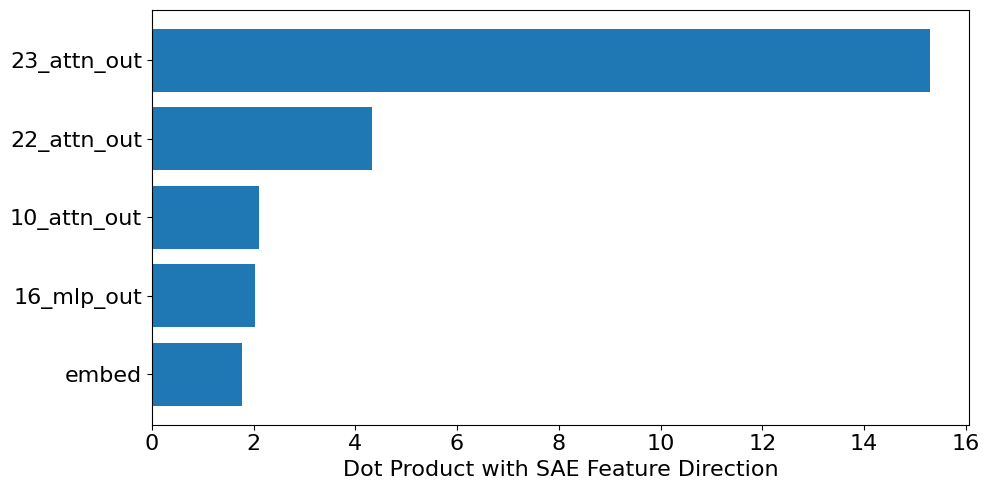}
        \subcaption{Layer 23}
    \end{minipage}
    \hfill
    \begin{minipage}[t]{0.48\linewidth}
        \centering
        \includegraphics[width=\linewidth]{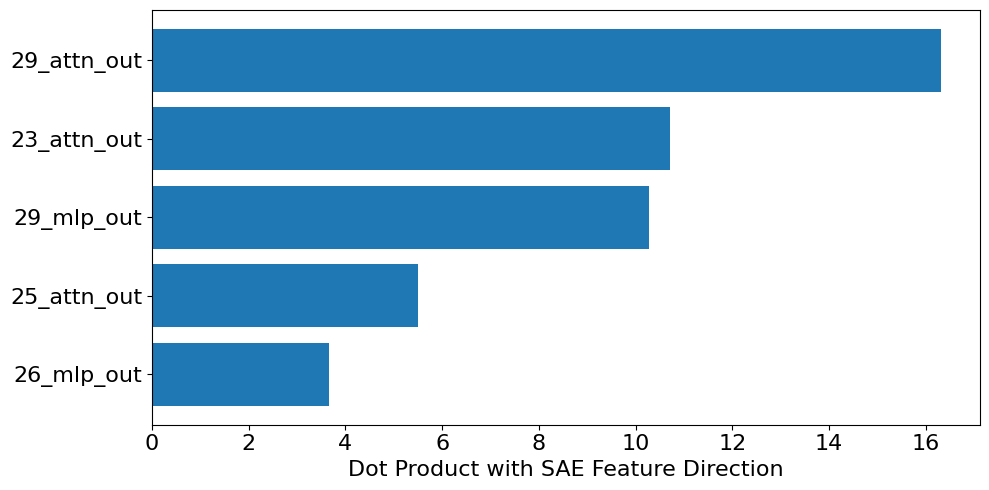}
        \subcaption{Layer 29}
    \end{minipage}
    \caption{
    \textbf{Decomposition of the residual stream in Gemma-2-9B showing the top contributors to the Japanese SAE feature direction at Layers 23 and 29.} Each bar represents the dot product between a residual component (MLP or attention output from a previous layer) and the language feature direction identified at the current layer. At Layer 23, attention heads in the same layer dominate, suggesting localized construction. At Layer 29, strong contributions also come from within the same layer, but with notable support from Layer 23 and earlier MLPs.
    }
\end{figure}
\FloatBarrier

\subsubsection{Spanish}
\begin{figure}[H]
    \centering
    \begin{minipage}[t]{0.48\linewidth}
        \centering
        \includegraphics[width=\linewidth]{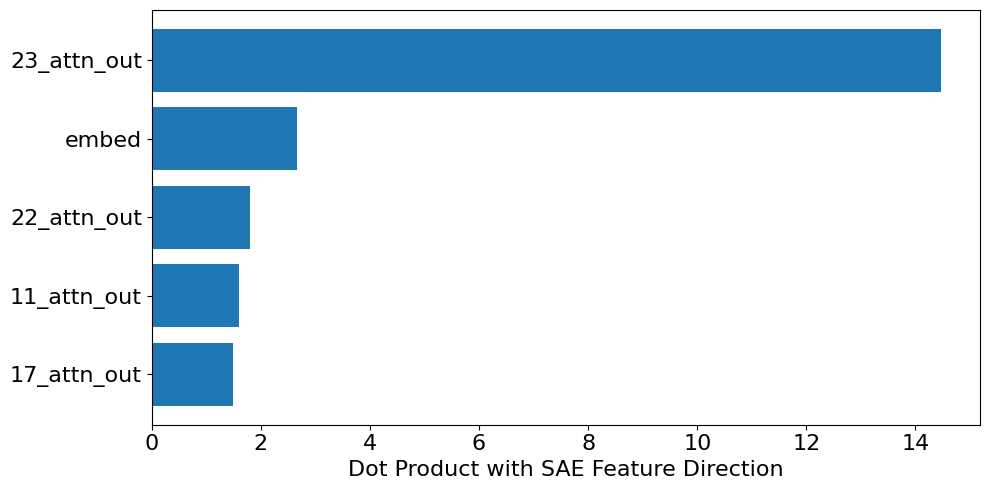}
        \subcaption{Layer 23}
    \end{minipage}
    \hfill
    \begin{minipage}[t]{0.48\linewidth}
        \centering
        \includegraphics[width=\linewidth]{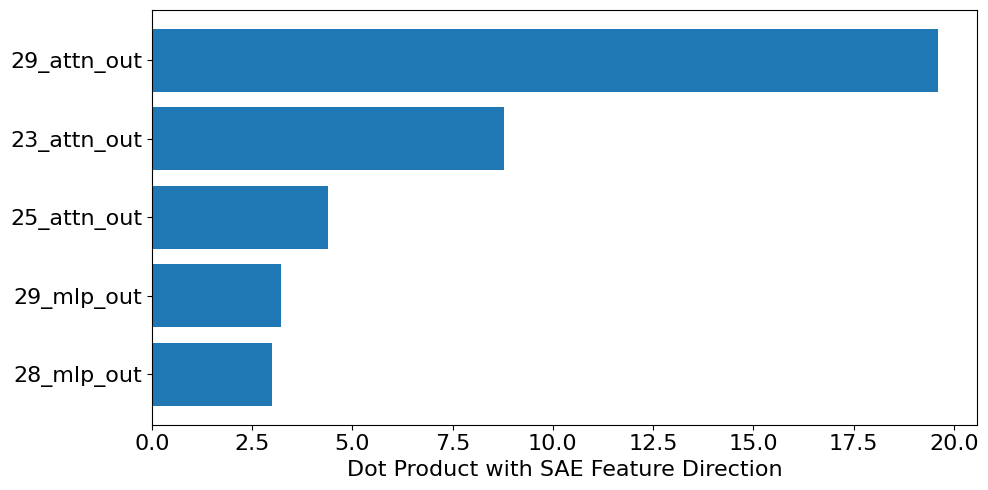}
        \subcaption{Layer 29}
    \end{minipage}
    \caption{
    \textbf{Decomposition of the residual stream in Gemma-2-9B showing the top contributors to the Spanish SAE feature direction at Layers 23 and 29.} Each bar represents the dot product between a residual component (MLP or attention output from a previous layer) and the language feature direction identified at the current layer. At Layer 23, attention heads in the same layer dominate, suggesting localized construction. At Layer 29, strong contributions also come from within the same layer, but with notable support from Layer 23 and earlier MLPs.
    }
\end{figure}
\FloatBarrier

\subsubsection{French}
\begin{figure}[H]
    \centering
    \begin{minipage}[t]{0.48\linewidth}
        \centering
        \includegraphics[width=\linewidth]{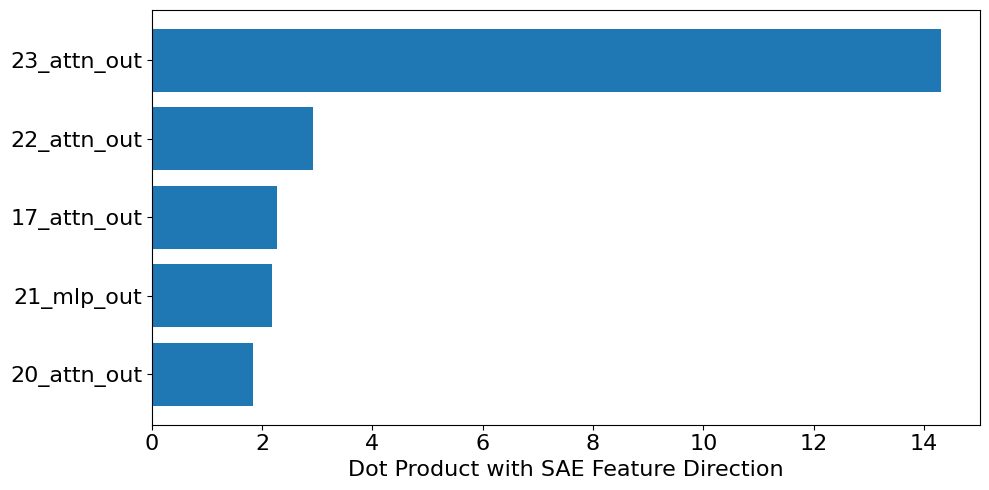}
        \subcaption{Layer 23}
    \end{minipage}
    \hfill
    \begin{minipage}[t]{0.48\linewidth}
        \centering
        \includegraphics[width=\linewidth]{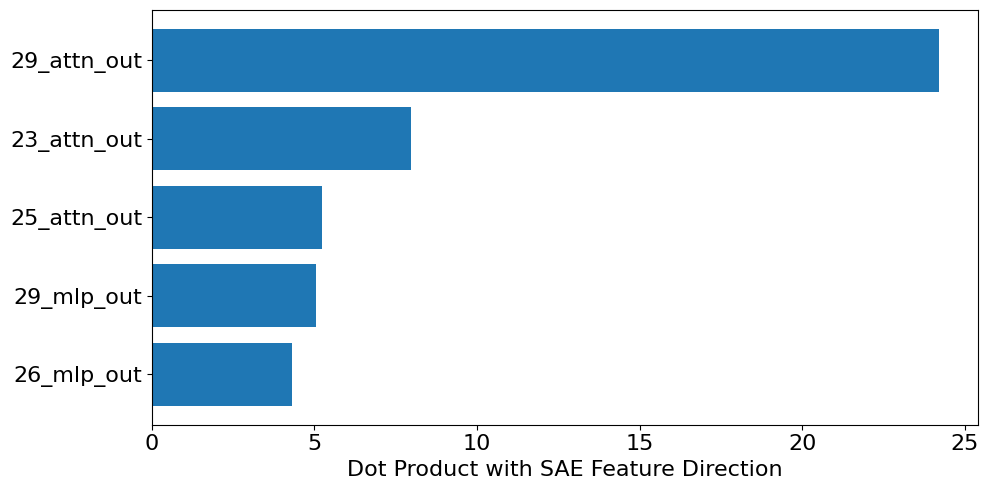}
        \subcaption{Layer 29}
    \end{minipage}
    \caption{
    \textbf{Decomposition of the residual stream in Gemma-2-9B showing the top contributors to the French SAE feature direction at Layers 23 and 29.} Each bar represents the dot product between a residual component (MLP or attention output from a previous layer) and the language feature direction identified at the current layer. At Layer 23, attention heads in the same layer dominate, suggesting localized construction. At Layer 29, strong contributions also come from within the same layer, but with notable support from Layer 23 and earlier MLPs.
    }
\end{figure}
\FloatBarrier

\subsection{Feature Persistence in Layers 30–33}
\label{sec:appendix-decomp-30-33}

We analyze the residual stream decomposition from Layers 30 through 33 for all four languages. These layers exhibit high steering scores for one or more languages, but no dominant localized contributors, suggesting inherited representations.

\subsubsection{Chinese}
\begin{figure}[H]
  \centering
  \begin{minipage}[t]{0.49\linewidth}
    \centering
    \includegraphics[width=\linewidth]{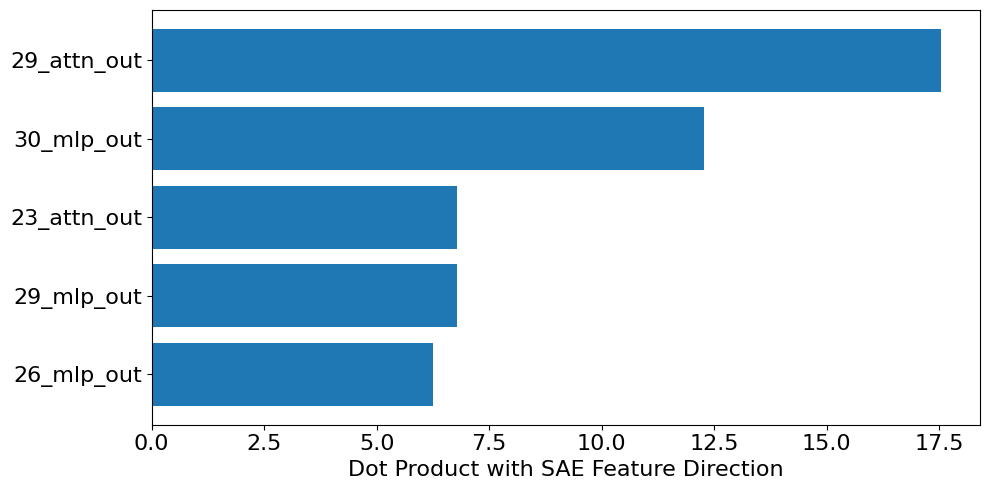}
    \subcaption{Layer 30}
  \end{minipage}%
  \hfill
  \begin{minipage}[t]{0.49\linewidth}
    \centering
    \includegraphics[width=\linewidth]{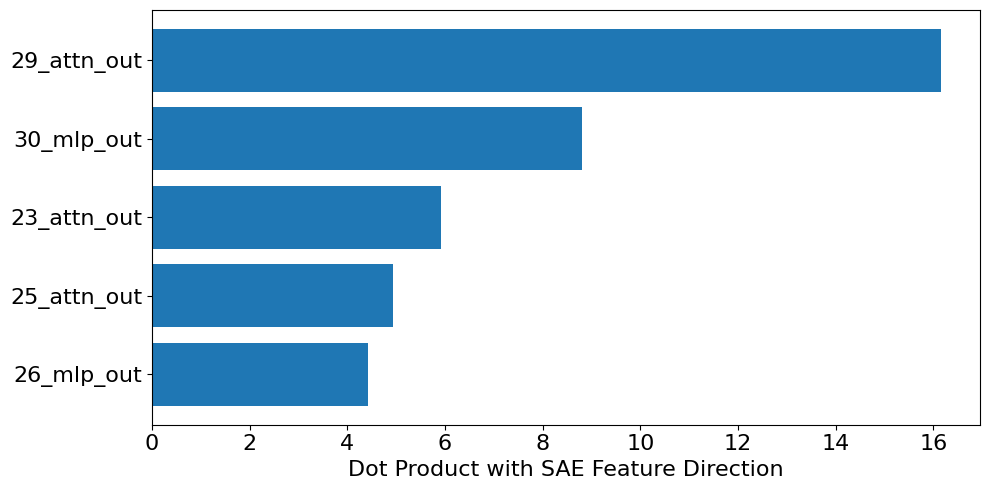}
    \subcaption{Layer 31}
  \end{minipage}
  \begin{minipage}[t]{0.49\linewidth}
    \centering
    \includegraphics[width=\linewidth]{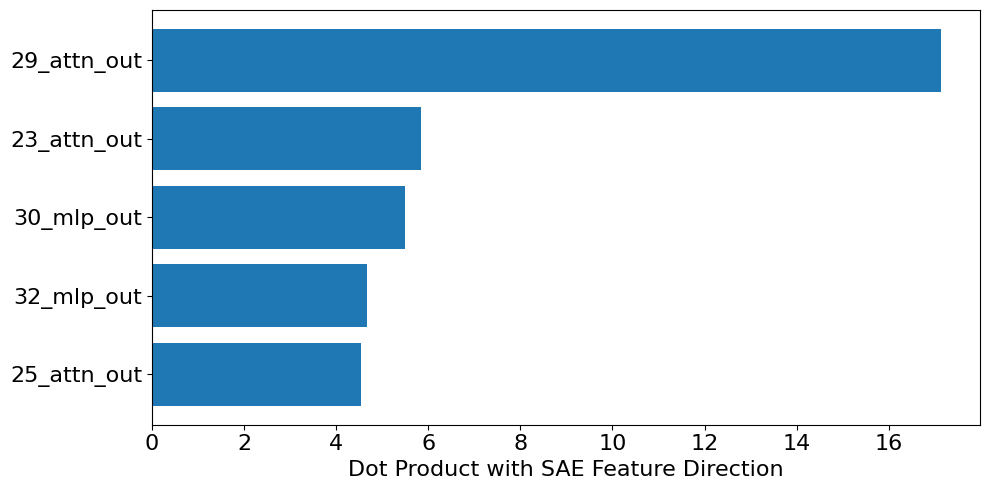}
    \subcaption{Layer 32}
  \end{minipage}%
  \hfill
  \begin{minipage}[t]{0.49\linewidth}
    \centering
    \includegraphics[width=\linewidth]{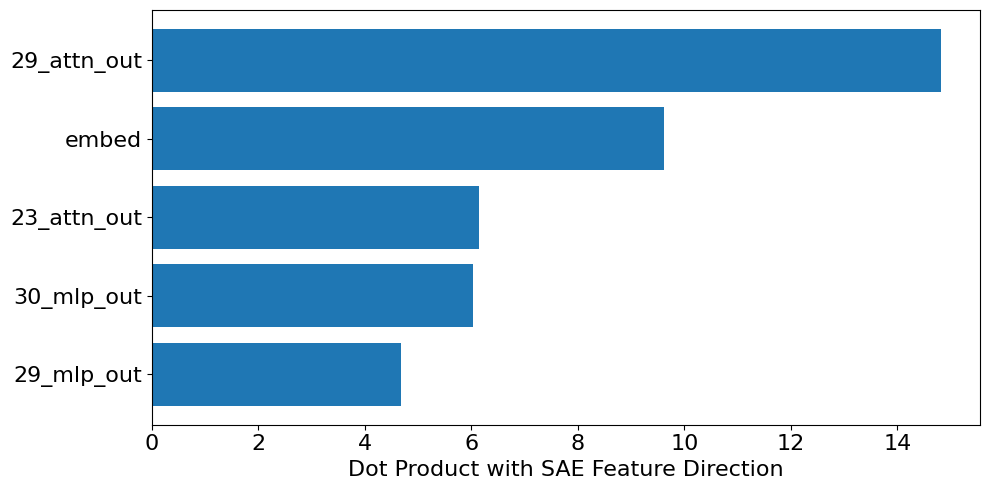}
    \subcaption{Layer 33}
  \end{minipage}
  \caption{Top 5 contributors to the Chinese SAE feature direction in layers 30 to 33, in incrementing order. Each bar represents the dot product between a residual component (MLP or attention output from a previous layer) and the language feature direction identified at the corresponding layer.}
\end{figure}
\FloatBarrier

\subsubsection{Japanese}
\begin{figure}[H]
  \centering
  \begin{minipage}[t]{0.49\linewidth}
    \centering
    \includegraphics[width=\linewidth]{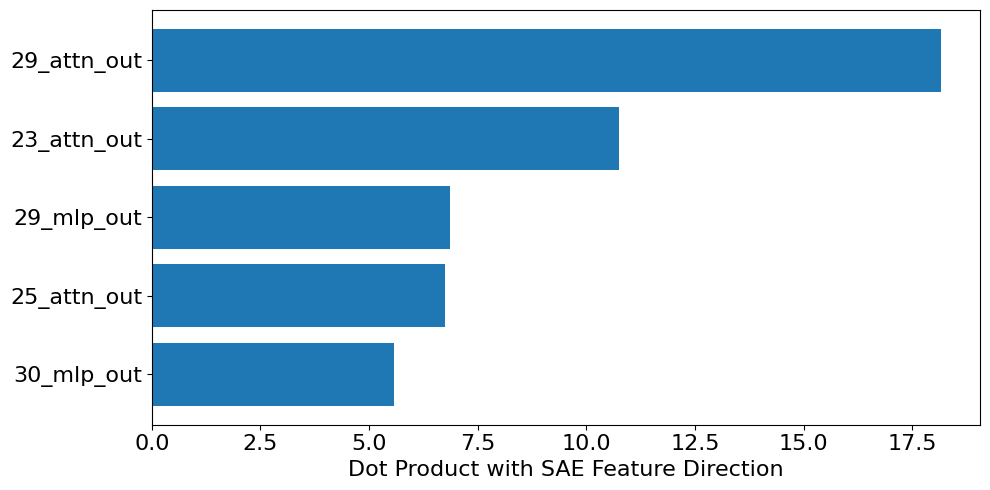}
    \subcaption{Layer 30}
  \end{minipage}%
  \hfill
  \begin{minipage}[t]{0.49\linewidth}
    \centering
    \includegraphics[width=\linewidth]{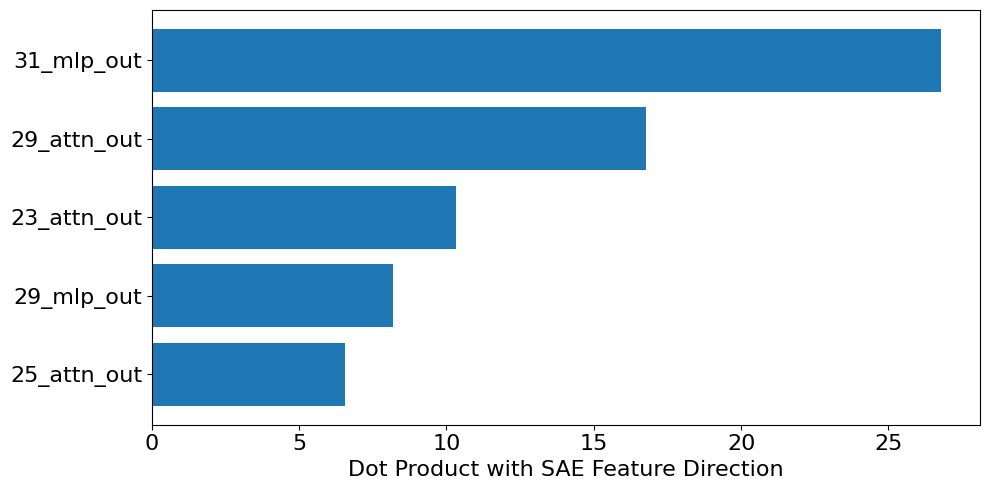}
    \subcaption{Layer 31}
  \end{minipage}
  \begin{minipage}[t]{0.49\linewidth}
    \centering
    \includegraphics[width=\linewidth]{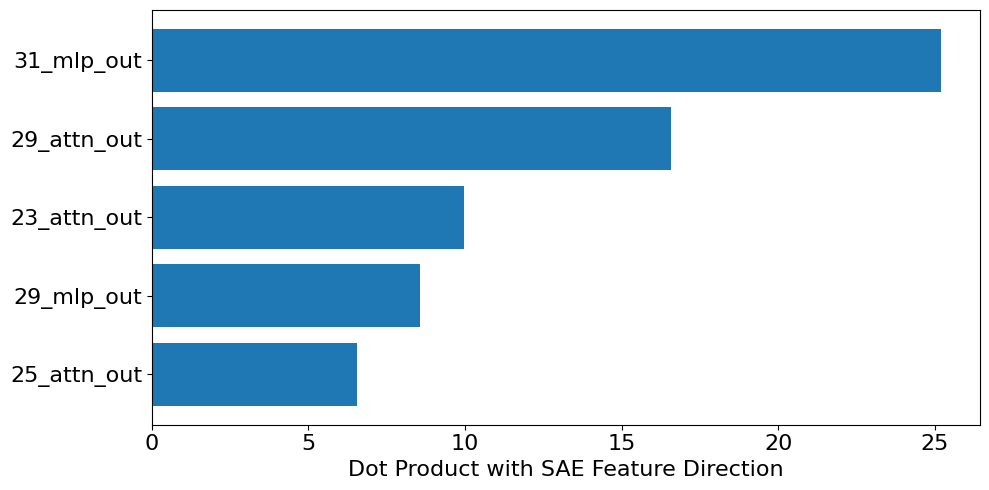}
    \subcaption{Layer 32}
  \end{minipage}%
  \hfill
  \begin{minipage}[t]{0.49\linewidth}
    \centering
    \includegraphics[width=\linewidth]{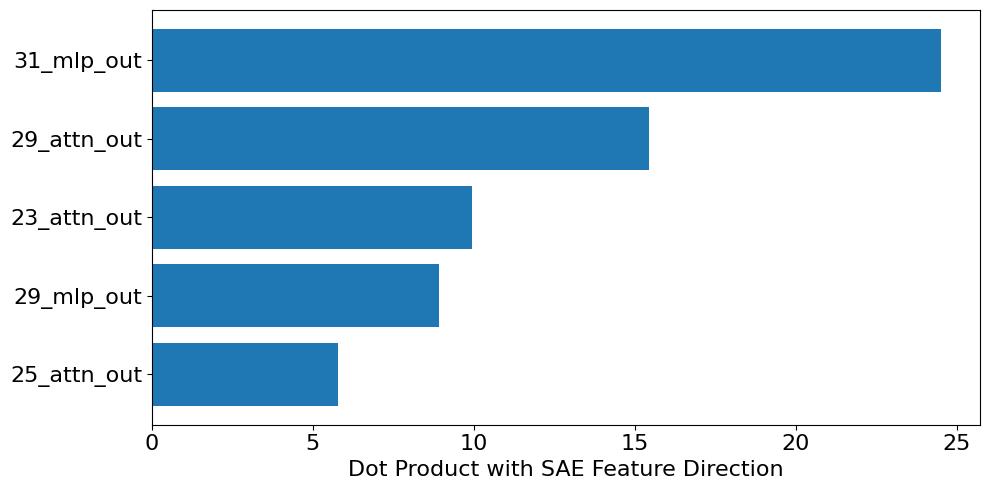}
    \subcaption{Layer 33}
  \end{minipage}
  \caption{Top 5 contributors to the Japanese SAE feature direction in layers 30 to 33, in incrementing order. Each bar represents the dot product between a residual component (MLP or attention output from a previous layer) and the language feature direction identified at the corresponding layer.}
\end{figure}
\FloatBarrier

\subsubsection{Spanish}
\begin{figure}[H]
  \centering
  \begin{minipage}[t]{0.49\linewidth}
    \centering
    \includegraphics[width=\linewidth]{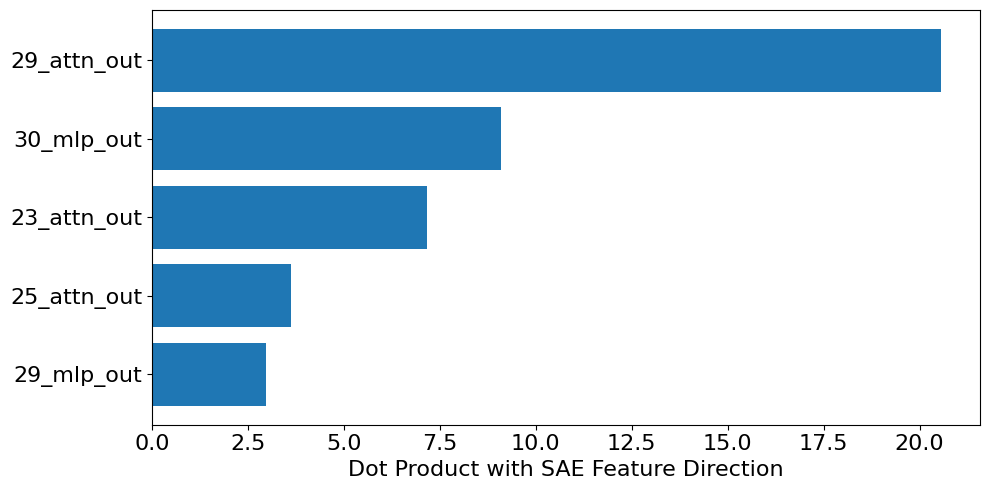}
    \subcaption{Layer 30}
  \end{minipage}%
  \hfill
  \begin{minipage}[t]{0.49\linewidth}
    \centering
    \includegraphics[width=\linewidth]{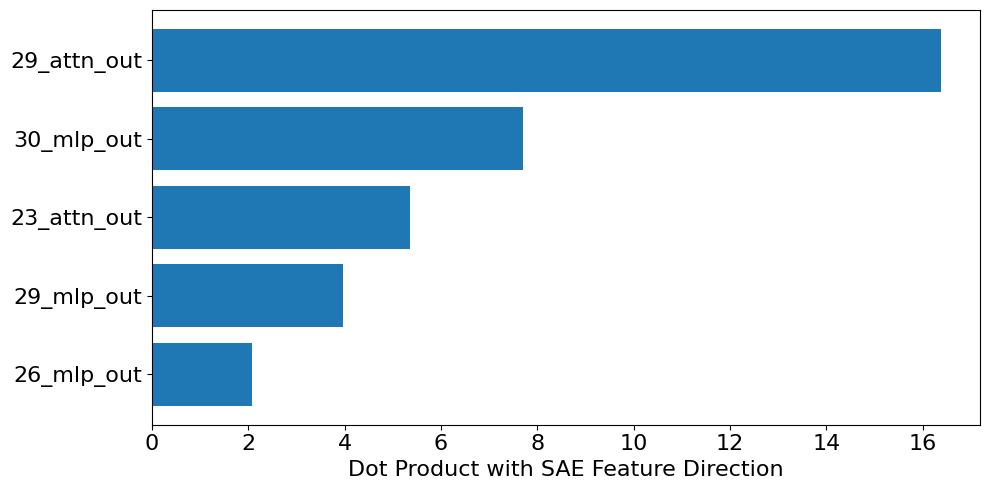}
    \subcaption{Layer 31}
  \end{minipage}
  \begin{minipage}[t]{0.49\linewidth}
    \centering
    \includegraphics[width=\linewidth]{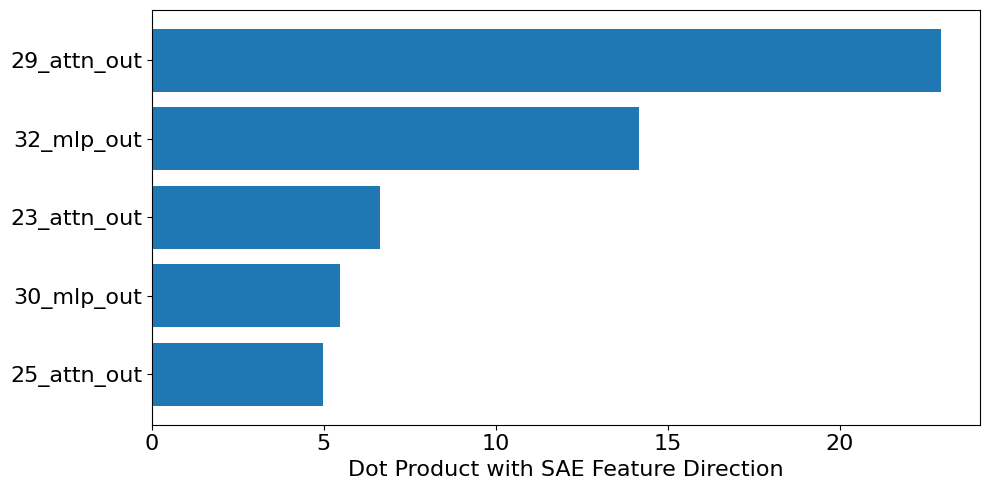}
    \subcaption{Layer 32}
  \end{minipage}%
  \hfill
  \begin{minipage}[t]{0.49\linewidth}
    \centering
    \includegraphics[width=\linewidth]{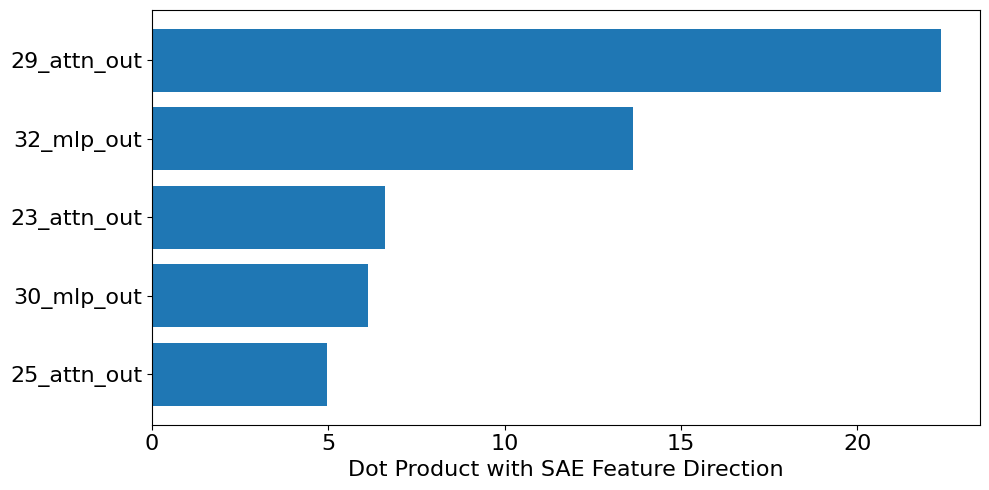}
    \subcaption{Layer 33}
  \end{minipage}
  \caption{Top 5 contributors to the Spanish SAE feature direction in layers 30 to 33, in incrementing order. Each bar represents the dot product between a residual component (MLP or attention output from a previous layer) and the language feature direction identified at the corresponding layer.}
\end{figure}
\FloatBarrier

\subsubsection{French}
\begin{figure}[H]
  \centering
  \begin{minipage}[t]{0.49\linewidth}
    \centering
    \includegraphics[width=\linewidth]{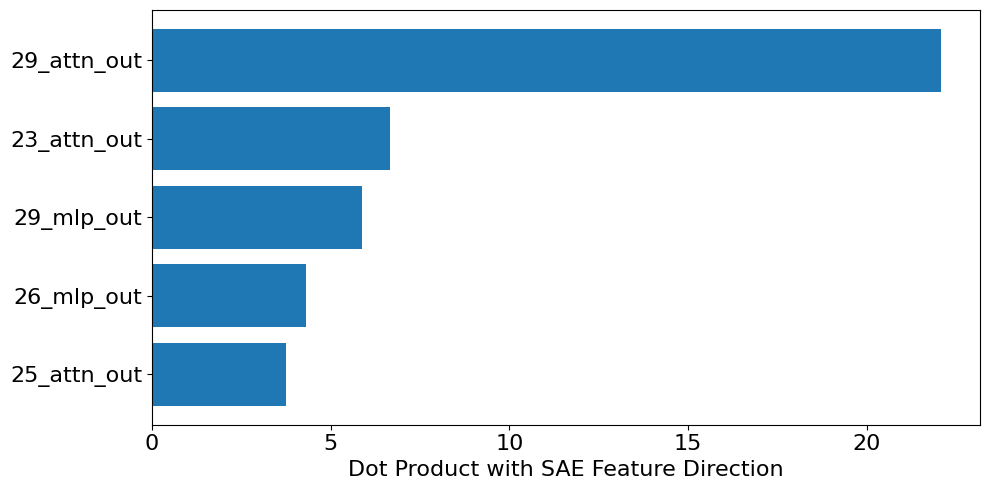}
    \subcaption{Layer 30}
  \end{minipage}%
  \hfill
  \begin{minipage}[t]{0.49\linewidth}
    \centering
    \includegraphics[width=\linewidth]{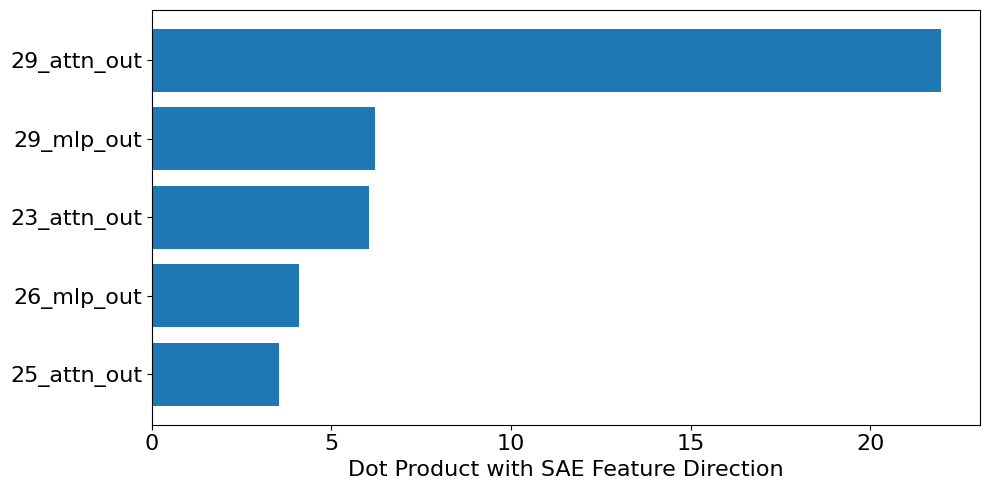}
    \subcaption{Layer 31}
  \end{minipage}
  \begin{minipage}[t]{0.49\linewidth}
    \centering
    \includegraphics[width=\linewidth]{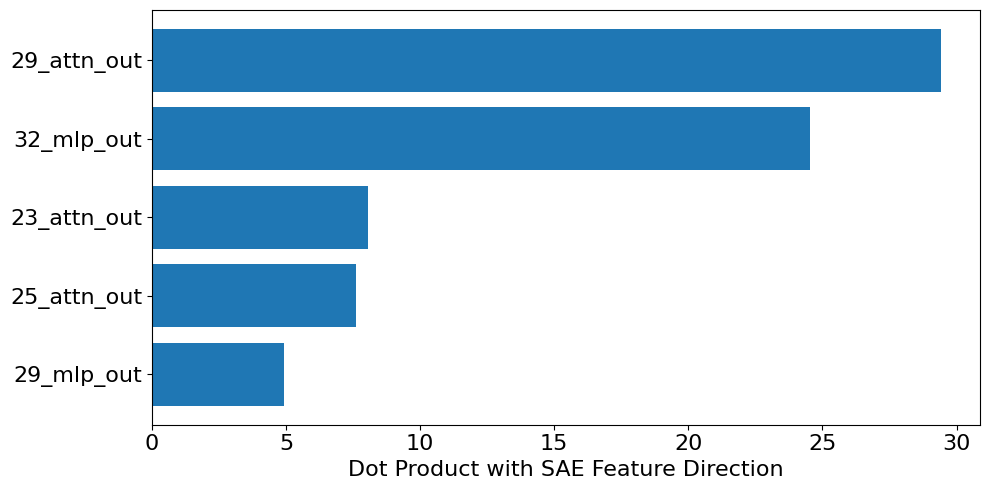}
    \subcaption{Layer 32}
  \end{minipage}%
  \hfill
  \begin{minipage}[t]{0.49\linewidth}
    \centering
    \includegraphics[width=\linewidth]{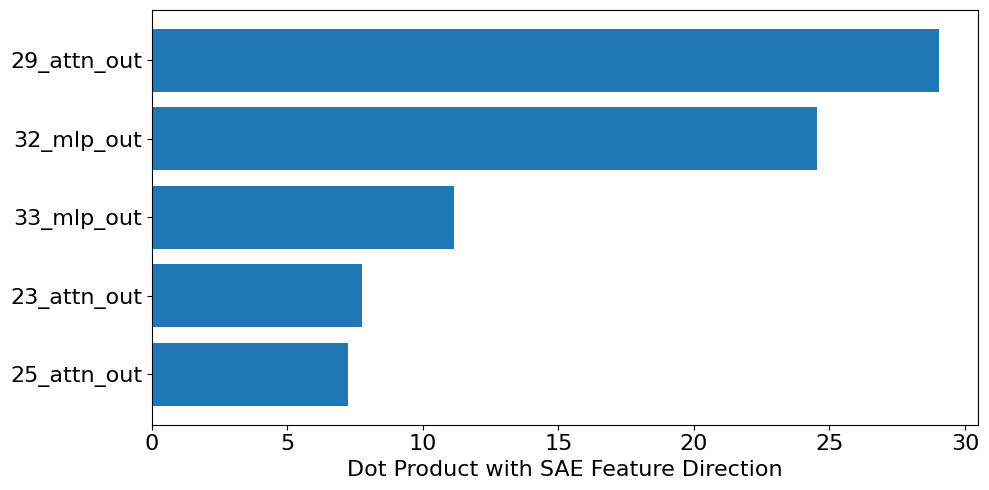}
    \subcaption{Layer 33}
  \end{minipage}
  \caption{Top 5 contributors to the French SAE feature direction in layers 30 to 33, in incrementing order. Each bar represents the dot product between a residual component (MLP or attention output from a previous layer) and the language feature direction identified at the corresponding layer.}
\end{figure}
\FloatBarrier

\end{document}